\documentclass[11pt, a4paper]{article}

\usepackage[margin=2.5cm]{geometry}

\usepackage{authblk}

\usepackage{graphicx}
\usepackage{multirow}
\usepackage{amsmath,amssymb,amsfonts}
\usepackage{amsthm}
\usepackage[scr=rsfso]{mathalfa}
\usepackage[title]{appendix}
\usepackage{xcolor}
\usepackage{textcomp}
\usepackage{booktabs}
\usepackage{algorithm}
\usepackage{algorithmicx}
\usepackage{algpseudocode}
\usepackage{listings}
\usepackage{physics}
\usepackage{subcaption}
\usepackage{hyperref}
\usepackage{natbib}

\theoremstyle{plain}

\theoremstyle{definition}

\newtheorem*{defn}{Definition}

\theoremstyle{remark}

\DeclareMathAlphabet{\mathbbold}{U}{bbold}{m}{n}

\newcommand{\keywords}[1]{\par\noindent\textbf{Keywords:} #1}

\begin{document}

\title{Constituency Optimisation Through Hamiltonian Representation Of Mandates (COTHROM): Algorithmic Redistricting of Irish Election Boundaries}

\author[1,2]{Ruaidhr\'i Campion\thanks{campioru@tcd.ie}}
\author[1]{Matthew Fenlon}
\author[1]{Joshua Cooney Mercadal}
\author[1]{Casey Farren-Colloty}
\author[4]{Eliza Somerville}
\author[1,3]{Michael A.J. Mitchell\thanks{mitchemi@tcd.ie}}

\affil[1]{The Problem Solving Association C.L.G.}
\affil[2]{School of Mathematics, Trinity College Dublin}
\affil[3]{School of Physics, Trinity College Dublin}
\affil[4]{School of Engineering, Trinity College Dublin}

\date{}

\maketitle

\begin{abstract}
Electoral redistricting in Ireland's Proportional Representation Single Transferable Vote (PR-STV) system faces the challenge of selecting an optimally representative set of electoral boundaries from an enormous set of possible configurations, and where ``representative'' is a delicate balance of constitutional objectives that are often in tension with one another.
We present the first computational framework for Irish electoral redistricting that systematically optimises across multiple constitutional requirements while making trade-offs explicit and quantifiable.
The electoral redistricting problem is parsed using statistical physics, where constitutional objectives are considered as terms in a Potts Hamiltonian. Markov Chain Monte Carlo (MCMC) methods and simulated annealing are employed to minimise this objective function, systematically exploring this configuration space, with coupling constants as proxies for objective weightings. Multi Criterion Decision Analysis (MCDA) and Pareto Optimality is then utilised to remedy the ambiguity in choosing a certain objective weighting combination over others. With respect to proportional representation and compactness objectives evaluated in County Cork, COTHROM consistently improves on the existing legal constituency boundaries for a range of objective weightings.
\end{abstract}


\vspace{1em}


\section{Introduction}
In Ireland, the boundaries of the Dáil constituencies are redrawn at least once every twelve years to reflect changes in the distribution of the population. The rules governing this process are codified in Article 16.2 of the Constitution and Section 57 of the Electoral Reform Act 2022: between 171 and 181 representatives must be returned across constituencies of three, four, or five seats; constituencies must be contiguous; the breaching of county boundaries should be avoided ``as far as practicable''; and the ratio of representatives to population should, ``so far as it is practicable'', be the same throughout the country. The rules themselves are clearly stated, however, the legal text does not specify how to satisfy  simultaneously when they conflict with one another, which is left to Ireland's Coimisiún Toghcháin (Electoral Commission, referrred to as the ``EC'' from here on).

This is not only a question of legal interpretation but also one of combinatorial structure. The smallest indivisible units of the partition, known as the Electoral Divisions (EDs), form a graph in which each constituency is a connected subgraph, and the constitutional and statutory rules carve out a very large subset of admissible partitions. Within that admissible set, the soft objectives induce a preference order: a partition that breaches few county boundaries may necessarily admit larger departures from proportional representation, and a partition that achieves tight proportional representation may require many such breaches. These tensions are not accidents of any particular plan, but are instead properties of the constrained graph itself. They cannot be read off of the legal text, and they are not made transparent by any single proposed configuration of boundaries. 

In practice, Irish constituency boundaries are drawn manually, guided by legal interpretation and public consultation \cite{workingpapers2023}. This process offers no systematic basis for comparison: once the EC judges a configuration to be compliant with the statutory terms of reference, neither the Commission nor a subsequent judicial review is required to demonstrate that no admissible alternative satisfies those terms more effectively. The gap is methodological rather than legal, since the law requires a permissible plan, rather than a near-optimal one. Framing the redistricting task as an optimisation problem makes this gap visible. By encoding the constitutional and statutory criteria as objective functions and exploring the admissible space algorithmically, we can quantify how different configurations trade off against one another and characterise where the law's softer commitments (i.e. ``as far as practicable'', ``regard to geographic considerations'') leave room for substantively different choices. 

\subsection{Redistricting in Ireland}
Ireland's bicameral parliament is known as the Oireachtas, and consists of two houses: the Dáil Éireann (lower house) and the Seanad Éireann (upper house)~\cite{ParliamentIreland1937}.
Members of the Dáil, known as Teachtaí Dála (TDs), are elected from multi-seat electoral areas called constituencies through a Proportional Representation system using the Single Transferable Vote (PR-STV)~\cite{McBride01121996}. 
Ireland's PR-STV is an uncommon method of election shared most notably by the House of Representatives of Malta~\cite{MaltaConstitution1964Art52} 
and the Australian Senate~\cite{FarrellMcAllister2006}, decreed in Article 16.2.5\textdegree of the Irish Constitution~\cite{ParliamentIreland1937Art16}. 
The STV system means that during an election, multiple representatives will be elected to a given constituency region. Unlike a first-past-the-post system, a voter will number their ballot with their favoured candidates in order of preference, so that when a candidate receives a predetermined threshold of votes to be deemed elected (the ``Droop'' quota~\cite{Droop_1881}), their surplus votes are distributed amongst the other candidates in keeping with their next-ranked choice. There are different methods of determining the surplus, such as the Gregory method~\cite{prsa_greggregory} 
or the whole-vote method~\cite{10.1093/0199257566.001.0001}, 
although it is not relevant to our work. The lowest-voted candidates also have their votes redistributed in turn, and this process continues until either all votes have been distributed, or the allocated number of representatives for that region have been elected to office. 

The compelling aspect of this voting system lies in its ability to reflect the underlying distribution of political sentiments by having more representatives elected per region, minimising ``rounding error'', so to speak. This is enshrined in Article 16.2.6\textdegree of the Irish Constitution:

\begin{quote}
\textbf{Article 16.2.6\textdegree} \textit{No law shall be enacted whereby the number of members to be returned for any constituency shall be less than three.}
\end{quote}

This is coupled with a principle of Proportional Representation, which is conveyed in Article 16.2.3\textdegree:

\begin{quote}
\textbf{Article 16.2.3\textdegree} \textit{The ratio between the number of members to be elected at any time for each constituency and the population of each constituency, as ascertained at the last preceding census, shall, so far as it is practicable, be the same throughout the country.}
\end{quote}

In order to facilitate proportional representation as population density changes in Ireland, the boundaries must be redrawn, which is captured in Article 16.2.4\textdegree:
\begin{quote}
\textbf{Article 16.2.4\textdegree} \textit{The Oireachtas shall revise the constituencies at least once in every twelve years, with due regard to changes in distribution of the population...}
\end{quote}

Since 2023, this redistricting responsibility has been vested in the EC, which is an independent body which replaced Ireland's previous ad-hoc Constituency Commissions~\cite{ElectoralReformAct572022}. There are additional constitutional constraints on this redistricting process, codified in the Electoral reform act Section 57~\cite{ElectoralReformAct572022}, which are summarised as 
\begin{itemize}
    \item \textbf{Section 57.2 (a)}: There must be at least 171 to at most 181 representatives.
    \item \textbf{Section 57.2 (b)}: Constituencies can have 3, 4, or 5 representatives.
    \item \textbf{Section 57.2 (c)}: ``the breaching of county boundaries shall be avoided as far as practicable''.
    \item \textbf{Section 57.2 (d)}: Constituencies must be contiguous.
    \item \textbf{Section 57.2 (e)}: ``there shall be regard to geographic considerations including significant physical features and the extent of and the density of population in each constituency''
    \item \textbf{Section 57.2 (f)}: The EC should endeavour to maintain the temporal stability of the boundaries, subject to the above items. 
\end{itemize}

The EC refers to this set of legal mandates as the ``statutory terms of reference'' or ``terms of reference''. There are additional, implicit considerations in the drawing of electoral boundaries. Boundaries have historically been drawn following a public consultation~\cite{AnCoimisiunToghchain2023}, incorporating input from both the public and elected representatives. Historically, elected representatives have advocated for certain boundaries on the basis of consolidating where they expend their political effort~\cite{Coakley2008}, and in the most recent Constituency Review Report of 2023, the public submissions were largely to do with the breaching of county boundaries and their perceived impact on effective representation~\cite{AnCoimisiunToghchain2023}. This brings us to an important limitation of our attempt to tackle this problem algorithmically - the EC faces implicit constraints stemming from the public consultation which we do not endeavour to model here. Our work represents an algorithmic optimum under the explicitly stated constitutional and legal objectives. While not implemented in this work, we do suggest a means by which these public consultation insights regarding certain community connections can be modelled explicitly, such that one can be sure they are enforced equitably. 

Ireland is an excellent case study for PR-STV algorithmic redistricting, since it is small enough to be tractable, complex enough to be interesting, and the process is subject to transparent legal objectives. 
Since the EC in Ireland is independent, the central problem of the redistricting task is not one of detecting partisan manipulation, but optimisation under constitutional constraints and primary legislation, and understanding the interplay between these legal objectives. Currently, the redistricting process is performed manually, which is not suited to strict reproducibility and boundary optimality, but presents an opportunity for this kind of analysis. 

\subsection{Related Work}
The process of making the redistricting task amenable to mathematical and computational optimisation has been studied from various academic perspectives since the 1960's, beginning with punch-card deterministic optimisation algorithms~\cite{nagel1965simplified,garfinkel1970optimal} and least-squares methods~\cite{weaver1963procedure}. For comprehensive surveys spanning the field's development, see Ricca et al.~\cite{ricca_et_al_2013} and the interdisciplinary volume by Duchin and Walch~\cite{duchin_walch_2022}.
Given this combinatorial enormity, the field has moved decisively towards Markov Chain Monte Carlo (MCMC) methods, which offer theoretical guarantees regarding convergence to a target distribution and provide a principled framework for exploring the vast space of possible configurations. 

Our introduction to this problem was through the work of Chou and Li~\cite{chou_li_2006,chou_li_2007_jmmm}, in which the redistricting problem is parsed using statistical physics: the system is ushered towards a ground state via MCMC, and the components of the energy function are configured to each admit a minimum at fulfilment of their respective legal constraints or objectives. Ultimately, it is this line of enquiry that inspired us to apply it to Ireland's particular constitutional objectives and PR-STV context. We similarly employ this Potts model~\cite{wu_1982} framing; a multi-spin magnetic system in which each spin value corresponds to a district assignment-partly because the physics analogy enables us to reason about the convergence process, identifying when ``phase transitions'' occur and their implications for the MCMC algorithm's ability to find an approximately optimal state.

While our approach and that of Chou and Li~\cite{chou_li_2006} rely on sampling from or optimising over a stationary distribution, not all approaches follow this paradigm. The ReCom (Recombination) family of Markov chains, introduced by DeFord, Duchin, and Solomon~\cite{deford_duchin_solomon_2021}, sidesteps the need for an explicit energy function and instead samples from a spanning-tree distribution on graph partitions. ReCom generates large ensembles of plausible maps which can then be used to characterise whether a proposed plan is a statistical outlier, a framing that has proven highly influential in US redistricting litigation. The associated open-source software, GerryChain~\cite{gerrychain_2018}, has been used in multiple court cases and by redistricting commissions, and constitutes the dominant practical toolkit in the field. A related sequential Monte Carlo approach by McCartan and Imai~\cite{mccartan_imai_2023} sidesteps MCMC mixing time concerns.

Other approaches to the redistricting problem include the graph-cut MCMC formulation of Fifield et al.~\cite{fifield_et_al_2020}, which incorporates simulated and parallel tempering to improve mixing; column generation heuristics~\cite{gurnee_shmoys_2021}; evolutionary algorithms~\cite{10.1007/978-3-540-72590-9_174}; and search-tree methods. Other works focus on the geometry of the problem, with Duchin and Tenner~\cite{duchin_tenner_2024} arguing for discrete compactness metrics such as cut edges over traditional contour-based scores like Polsby--Popper. The interaction of computational redistricting with the legal process and its practical uptake is discussed in Becker et al.~\cite{becker_et_al_2021}.

A distinguishing feature of our approach is the explicit use of multi-criteria decision analysis (MCDA) and Pareto front analysis, rather than collapsing all objectives into a single scalar energy function. In any redistricting problem, multiple competing objectives must be balanced, population equality, compactness, county boundary preservation, and the relative weighting of these objectives involves value judgements that are, fundamentally, normative rather than mathematical. By establishing the Pareto front, we make these trade-offs explicit and transparent, enabling decision-makers to select from among non-dominated solutions according to their own priorities.

The multi-objective optimisation literature provides the theoretical and algorithmic foundations for this approach. The concept of Pareto dominance and the algorithms for approximating Pareto fronts are well established, with Deb's NSGA-II~\cite{deb_et_al_2002} being the most widely used evolutionary multi-objective algorithm (see also~\cite{deb_2001} for a comprehensive treatment). In the redistricting context specifically, Swamy, King, and Jacobson~\cite{swamy_et_al_2023} present the closest comparator to our work: they formulate bi-objective politically fair districting problems using Mixed Integer Linear Programming (MILP), approximating the Pareto front via the $\varepsilon$-constraint method. However, their approach differs from ours in several important respects: it targets single-member plurality districts rather than multi-seat STV constituencies, it employs MILP rather than a Potts model framing.

\subsection{Problem Formulation}
As we will see in detail in the next section, the legal requirements are each given a mathematical translation, say $H_\alpha(\sigma)$ for requirement $i$ evaluated the configuration ($\sigma)$, which are then combined in a weighted sum to form a cost function, $H(\sigma) = \sum_\alpha J_\alpha H_\alpha(\sigma)$ with weights $\{J_\alpha\}$. Once stated in this form, the redistricting problem decomposes into two questions of quite different character. 

The first is computational: given a set of coupling constants $\{J_\alpha\}$, find a configuration $\sigma$ that approximately minimises $H$. This is a discrete optimisation problem, and we address it using MCMC methods with a simulated annealing schedule, exploiting the physics interpretation of the Hamiltonian to reason about the minimisation landscape and diagnose convergence. 

The second question is normative: which $\{J_\alpha\}$ should be used? The Constitution and the Electoral Reform Act state the soft objectives but do not rank them, and provide no procedure for arbitrating between configurations that satisfy the hard constraints but differ in how aggressively they pursue each soft objective. Rather than committing to a single weighting, we treat $\{J_\alpha\}$ as a degree of freedom to be analysed: multi-criterion decision analysis (MCDA) is used to characterise the Pareto front of non-dominated configurations and identify where the soft objectives most sharply compete.

This decomposition surfaces a question that is, to our knowledge, not addressed explicitly in Irish redistricting practice. If the configuration that minimises the worst aggregate departure from the soft objectives requires those objectives to be weighted unequally, then the legal framework stating them in formally equal terms is silent on whether that unequal weighting is itself constitutionally permissible.

This paper is organised as follows. In the methods section, the representation of the problem and the constitutional objectives are detailed mathematically. We detail the convergence process, introducing the specific MCMC framework and simulated annealing scheduling used. The MCDA framework is discussed in greater detail in relation to the balance between compactness and proportional representation, Pareto optimality is discussed. In the results section, we analyse our findings and their implications. We validate our approach by comparing to the 2024 Commission boundaries, and examine the Pareto front from the MCDA study. Finally, we conclude.
\section{Methodology}
\subsection{Data and Metrics}\label{subsec:data-and-metrics}

The analysis was performed using Central Statistics Office (CSO) Electoral Divisions (EDs), using the 2022 statistical boundary files released by the CSO via Tailte \'Eireann~\cite{CSOPublishedEDs}. Although there are 3,440 legally-defined EDs, the CSO published statistics in 2022 for only 3,420, since certain low-population EDs must be amalgamated or have their boundaries amended to prevent statistical disclosure~\cite{CSO_Geographical_Classifications}. 
Each ED is additionally tagged with its parent administrative county.

The graph topology is determined by spatial adjacency: two EDs are connected if their boundary file geometries share a boundary. As a matter of convention, we consider island EDs to be connected to the nearest mainland ED with which they share an established ferry link.

Before introducing the objective function, we first discuss the quantitative metric that the EC uses for tracking proportional representation. We must first introduce a notion of a ``National Average'', which we denote $\langle P \rangle= P_T/S$, where $P_T$ is the total population of the Republic of Ireland, and $S$ is the total number of seats under consideration. The EC uses the following  metric,

\begin{equation}
    \label{eq:EC-variance}
    v_q \;=\; \frac{P_q/m_q - \langle P\rangle}{\langle P\rangle} 
              \,\times\, 100\%,
\end{equation}

where $q$ is the constituency being considered, $v_q$ is the referred to by the EC as the ``variance from the national average'', or simply ``variance'', and $m_q$ is the number of seats being considered for $q$.

\subsection{Optimisation Framework}\label{section: Optimisation Framework}
The next step is to encode the desired legal requirements into separate objective functions, such that their minimisation would enable the complete satisfaction of that objective. The main requirements for a configuration of constituencies are proportional representation ($P$), contiguity ($C$), compactness
($D$ from the Irish \textit{dlúthacht}), and county boundaries ($B$), represented by the Hamiltonians
\begin{align}\label{eq:Electoral Potts Hamiltonians}
    \begin{gathered}
        H_P = \sum_{q=1}^Q\left|\frac{P_q}{m_q\langle P\rangle}-1\right|, \\
        H_D = \sum_{\langle i,j\rangle}\left(1-\delta_{q_i,q_j}\right),
    \end{gathered} &&\begin{gathered}
        H_C = \sum_{q=1}^Q\left|G_q-1\right|, \\
        H_B = \sum_{i=1}^{N_{\mathrm{ED}}} \mathbf{1}\left[i \notin C(q_i)\right],
    \end{gathered}
\end{align}
where $P_q$ and $\langle P\rangle$ are the population of constituency $q$ and average population per constituency respectively, $m_q$ is the number of seats allocated to constituency $q$,\footnote{The Electoral Act of 1997 specifies $m_q = \{3,4,5\}$, i.e.~3-, 4-, and 5-seater constituencies.} $G_q$ is the number of contiguous groups in constituency $q$, $q_i$ is the constituency to which ED $i$ is allocated, and $C(q_i)$ is the set of indices demarcating the EDs belonging to the county in which the majority of $q_i$'s EDs belong to.

\begin{itemize}
    \item $H_P$ is minimised when each constituency represents its allocated integer multiple of the average population per constituency ($P_q = m_q \langle P\rangle$ for each $q=1,\ldots,Q$).
    
    \item $H_C$ is minimised when each constituency consists of exactly one connected group of EDs ($G_q=1$ for each $q=1,\ldots,Q$; note that the modulus $||$ is used so that a constituency $q$ contributes to $H_C$ even when no ED is assigned, i.e.~when $G_q=0$). In practice, a breadth-first search is performed to determine the number of groups into which a candidate constituency is split. An alternative, computationally cheaper contiguity algorithm for this problem is outlined in \autoref{EulerCharacteristic}.
    \item $H_D$: Compactness here is a discrete, cut-edge-style measure penalising shared inter-constituency boundaries. This has the effect of incentivising globular constituency shapes where minimised, rather than ``salamandering'' shapes. $H_D$ is minimised when all EDs are assigned to the same constituency ($q_i=q_j$ for all $i,j$, i.e.\ there are no inter-constituency boundaries).
    \item $H_B$: The constitutional language is ambiguous about what constitutes a county boundary breach. In the absence of an exact legal ruling we adopt a deliberately simple form: for each constituency we identify its primary county as the county containing the largest number of its constituent EDs and the county-boundary Hamiltonian counts the EDs in that constituency that lie outside the primary county. The term therefore scales with the number of off-primary-county EDs assigned to each constituency, but it makes no use of adjacency: an off-primary-county ED contributes the same penalty whether it borders the primary county or sits far from it. The original implementation of the county boundaries term utilised breach occurrence, although it did not perform well in terms producing configurations where county boundaries were respecting, and the current version performs much better. Additionally, we present an empirical analysis of past boundary breaches in \autoref{App:CountyBreaches}, however, validating the adoption of one functional form over another would require collaboration with the EC. It should be noted that previous Commissions were not required to avoid breaching provincial boundaries.
\end{itemize}


Each ``softer'' $H_\alpha$ (the objectives which are not to do with contiguity) is parametrised so that its penalty grows linearly with
the magnitude of the violation, i.e.\ a departure from proportional representation of between $5\%$ and $10\%$ under-representation contributes the same penalty as the between $10\%$ and $15\%$. Linear
penalties are a deliberately simple choice; they impose minimal
assumptions about whether small violations are disproportionately
worse than large ones, or vice versa. 
Quadratic or other non-linear parametrisations are admissible from a purely optimisation-theoretic
standpoint, but selecting one would amount to making a constitutional
interpretation that is outside the scope of this work. 

For $H_C$, we do depart from the minimal assumption - rather than classifying any amount for disconnectedness as equally ``bad'' (i.e. a constituency divided into $10$ disconnected regions is as legally invalid as a constituency split into $2$), we use a penalty which increases with the number of disconnected regions.

The final step is to form a linear combination of these Hamiltonians to create an objective function
\begin{equation}
    \label{eq:Electoral Potts Hamiltonian}H = \sum_{\alpha}J_\alpha H_\alpha,
\end{equation}
where $\alpha=P,C,D,\ldots$ refers to a requirement/guideline and the coupling constant $J_\alpha$ implicitly (and approximately) determines the consideration given to that requirement/guideline. Through this modified model, the redistricting problem has now been reduced to
\begin{itemize}
    \item translating the desired hierarchy of rules and guidelines $\left\{\alpha\right\}$ into numerical coupling constants $\left\{J_\alpha\right\}$, and
    \item determining the configuration of constituencies that minimises Eq.~\eqref{eq:Electoral Potts Hamiltonian} for a given set of constants $\left\{J_\alpha\right\}$.
\end{itemize}

It should be noted that we endeavoured to strictly enforce contiguity throughout the simulation, automatically rejecting proposals which violated it, but the mechanism interfered with the algorithm's convergence. Consequently, the contiguity term is treated like the other objective terms.

To make the coupling constants directly interpretable as proxies for
relative importance (\autoref{subsec:mcda}), we further normalise
each $H_\alpha$ to lie in $[0,1]$ by dividing by an upper bound of its
maximum:
\begin{align}
    \label{eq:Potts normalising factors}
    Z_P &= \frac{\sum_q m_q}{\min_q m_q} + Q-2, &
    Z_C &= N_{\mathrm{ED}}, &
    Z_D &= \sum_{\langle i,j\rangle}1, &
    Z_B &= N_{\mathrm{ED}} \frac{N_{\mathrm{county}}-1}{N_{\mathrm{county}}},
\end{align}
where $N_{\mathrm{ED}}$, $N_{\mathrm{county}}$ are the number of EDs and counties, respectively. The full objective function is then 
\begin{align}
    \label{eq:Normalised Electoral Potts Hamiltonian}
    H = \sum_{\alpha} J_\alpha \frac{H_\alpha}{Z_\alpha},
    \qquad \alpha \in \{P,\, C,\, D,\, B\},
\end{align}
with non-negative coupling constants $\{J_\alpha\}$ encoding the
relative weight assigned to each requirement. 

\subsection{The MCMC protocol}
\begin{figure}[!ht]
    \centering
    \includegraphics[width=\linewidth]{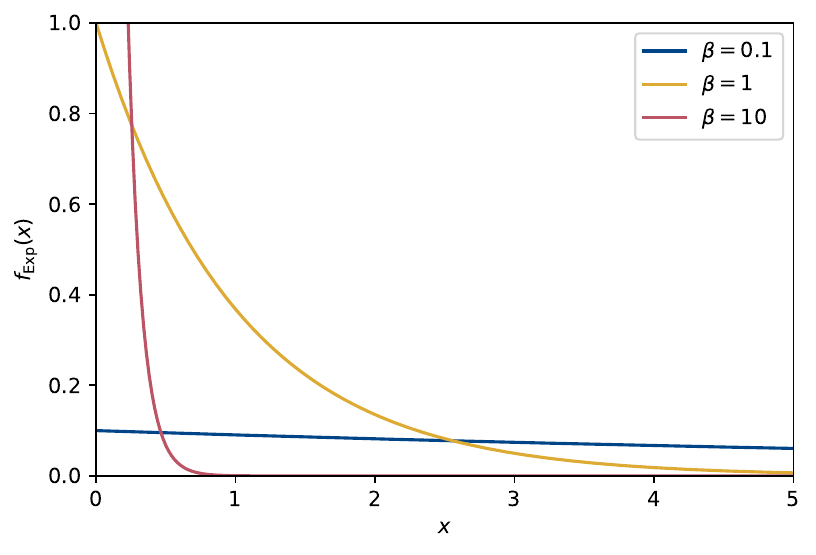}
    \caption{Exponential distribution (\autoref{eq:exp}) for $\beta=1,5,25$.}
    \label{fig:exp}
\end{figure}

In statistical mechanics, the (discrete) Boltzmann distribution\footnote{For simplicity, $k_B=1$ (or equivalently, $k_BT\to T$).}
\begin{equation}
    \label{eq:Boltzmann}\lambda(\sigma) \propto \exp\!\left(-\beta H(\sigma)\right)
\end{equation}
represents the probability of a system at temperature $T = 1/\beta$ being in a state $\sigma$ with energy $H(\sigma)\equiv H$. By comparison to the (continuous) exponential distribution
\begin{equation}
    \label{eq:exp}f_{\text{Exp}}(x) \propto e^{- \beta x},
\end{equation}
as depicted in \autoref{fig:exp}, as $\beta\to\infty$, $\lambda(\sigma)$ approaches a uniform distribution of the states with minimum energy. As such, if this distribution can be adequately sampled from, then doing so for a small temperature $\beta$ will return configurations that minimise $H$. 

Direct sampling from $\lambda(\sigma)$ is infeasible because the
partition function $Z(\beta) = \sum_\sigma \exp(-\beta H(\sigma))$ is
intractable for the configuration counts relevant here.\footnote{The
Stirling-number bound $S(3{,}420, Q) \gg 10^{5000}$ for $Q \geqslant 30$
rules out enumerative approaches. The restricted Stirling-number for considering larger constituencies does not change matters much in terms of incomprehensible scale.} We instead use Markov chain Monte
Carlo (MCMC) to construct a chain whose stationary distribution is
$\lambda$.

\subsection{Simulated Annealing}
\begin{figure}[!ht]
    \centering
    \begin{subfigure}[t]{0.3\linewidth}
        \includegraphics[width=\linewidth]{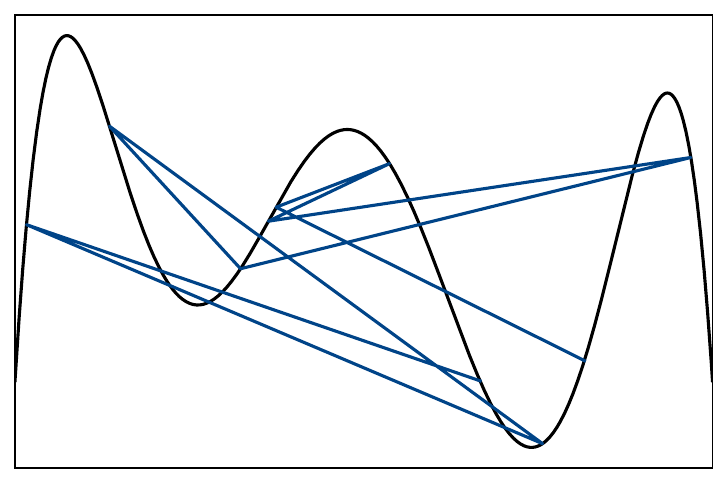}
        \caption{Fixed low $\beta$.}
        \label{subfig: high}
    \end{subfigure}
    \hfill
    \begin{subfigure}[t]{0.3\linewidth}
        \includegraphics[width=\linewidth]{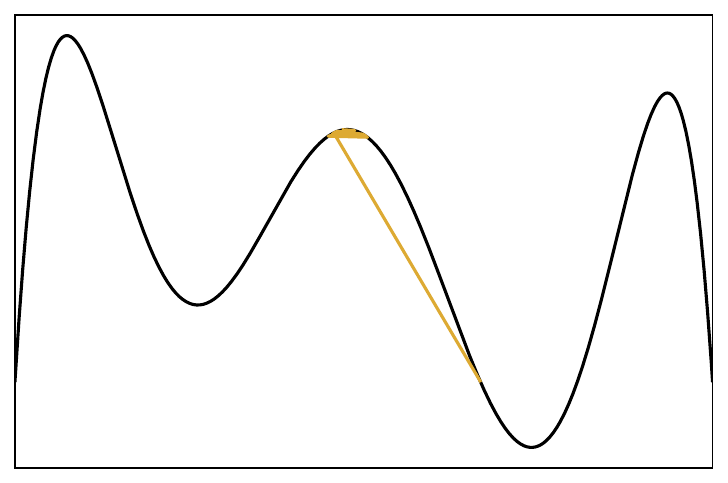}
        \caption{Fixed high $\beta$.}
        \label{subfig: low}
    \end{subfigure}
    \hfill
    \begin{subfigure}[t]{0.3\linewidth}
        \includegraphics[width=\linewidth]{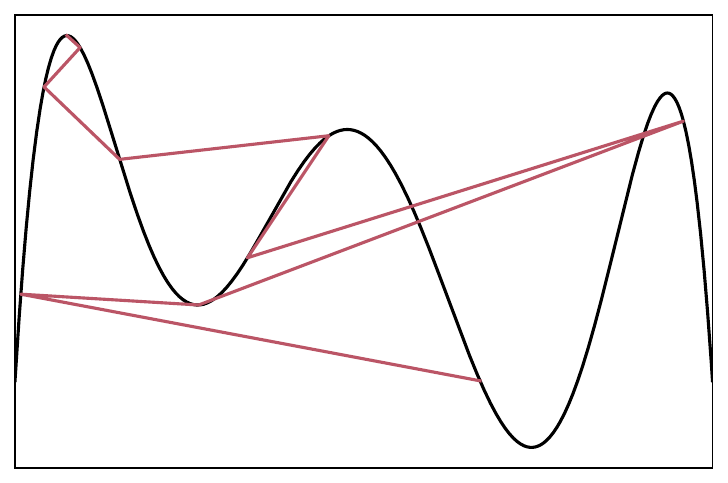}
        \caption{Increasing $\beta$.}
        \label{subfig: SA}
    \end{subfigure}
    \caption{A temperature-dependent algorithm maximising a
    one-dimensional function. At fixed low $\beta$ the algorithm samples
    broadly; at fixed high $\beta$ it becomes trapped near its initial
    state; with a decreasing schedule (simulated annealing) it tracks
    the equilibrium distribution from broad sampling at high $T$ down
    to concentration on the global maximum.}
    \label{fig: SA example}
\end{figure}

Fixing the optimisation to a single low temperature does not work in
practice. At low $\beta$ the chain samples essentially uniformly across
the entire configuration space (\autoref{subfig: high}); at high $\beta$ it
becomes trapped in whichever local minimum is closest to its initial
state (\autoref{subfig: low}). In either regime the probability of
recovering the global minimum from a random initial configuration is
vanishingly small. Simulated annealing~\cite{SA,Ross12} resolves this by
starting an annealing procedure at high $T$ and slowly reducing the temperature
according to a fixed schedule. Provided the annealing schedule is slow enough,
we enable the system to achieve equilibrium at each temperature, and as $\beta$
increases the distribution concentrates on the global minimisers
(\autoref{subfig: SA}). The name is borrowed from the materials
science technique of heating and slowly cooling a sample to relax it
into a low-energy atomic configuration.

We use a geometric schedule $T_k = \alpha^k T_0$ with $\alpha = 0.9$, and we anneal until no new configurations are proposed/there are no new accepted updates for an entire temperate run. The value $\alpha = 0.9$ was
chosen after preliminary tests showed that slower schedules produced no
detectable improvement in final ground-state energies but increased
runtime substantially. The initial temperature $T_0$ is set high enough that every possible proposal has at least a 99\% chance of being accepted at $T_0$ under the Metropolis criterion: 
\begin{equation}
    \label{eq:Potts init temp}
    \begin{gathered}
    \exp\!\left(-\frac{\Delta H_{\max}}{T_0}\right) \geqslant 0.99 \\
    \implies T_0 = 
    \frac{J_P\,\Delta H_{P,\max}
        + J_C\,\Delta H_{C,\max}
        + J_D\,\Delta H_{D,\max}
        + J_B\,\Delta H_{B,\max}}
         {-\ln 0.99},
    \end{gathered}
\end{equation}
where $\Delta H_{\alpha,\max}$ is an upper bound on the largest single-update change in
the corresponding component Hamiltonian $H_\alpha$.

\subsection{The Gibbs Sampler}
At each step the algorithm attempts to update the constituency
assignment of a single ED. Both the Metropolis algorithm and the Gibbs sampler are Metropolis-Hastings procedures and differ only in their proposal distribution: Metropolis proposes a candidate constituency uniformly at random, while the Gibbs sampler proposes from the conditional distribution over constituencies with the remainder of the configuration held fixed. We adopted the Gibbs sampler on the basis of preliminary testing, as it avoids the rejection inefficiency that the uniform-proposal Metropolis update can exhibit when many candidate moves are energetically unfavourable. The Gibbs step is more expensive per move, since it requires evaluating the energy of every candidate assignment for the chosen ED rather than a single proposal, but those evaluations are independent and parallelise straightforwardly. The Gibbs update for ED~$i$ is:
\begin{enumerate}
    \item For each candidate constituency $q\in\{1,\ldots,Q\}$,
    evaluate the energy change
    $\Delta H^{(q)} = H(\sigma\mid q_i \to q) - H(\sigma)$ (so that
    $\Delta H^{(q_i)} = 0$).
    \item Draw the new label $q_i'$ from the conditional distribution
    \begin{equation}
        \Pr(q_i' = q \mid \sigma_{\neq i})
        \;=\;
        \frac{\exp(-\Delta H^{(q)}/T)}
             {\sum_{q'=1}^{Q}\exp(-\Delta H^{(q')}/T)}.
    \end{equation}
    \item Set $q_i \leftarrow q_i'$.
\end{enumerate}
A full sweep applies the single-ED update to every ED in a fixed
arbitrary order. At each temperature the chain is run for a number of discarded sweeps before the temperature is  decremented, allowing the configuration to thermalise to the local equilibrium distribution at that temperature before cooling proceeds; any further sweeps performed at the same temperature are measurement sweeps used only for computing observables and do not affect the annealing trajectory. 

\subsection{Multi-Criterion Decision Analysis}\label{subsec:mcda}
The choice of coupling constants $\{J_\alpha\}$ in \autoref{eq:Normalised Electoral Potts Hamiltonian} is a normative question
that the constitutional and statutory framework does not resolve: the
soft objectives are stated but not numerically ranked. Rather than
committing to a single weighting, we use multi-criterion decision
analysis (MCDA) to characterise the structure of trade-offs between
the objectives.

The central object in MCDA is the Pareto front. Given a set of
candidate configurations and the vector of objective values
$(H_P, H_D, H_B)$ for each, configuration $\sigma_1$ is said to
\emph{dominate} configuration $\sigma_2$ if $\sigma_1$ is at least as good as $\sigma_2$ on
every objective and strictly better on at least
one~\cite{deb_2001}. A configuration is said to be Pareto
optimal, or equivalently ``non-dominated'', if no achievable
configuration dominates it; equivalently, improving any one objective
requires worsening at least one other. The set of all Pareto-optimal
configurations is the Pareto front, and represents the optimal
trade-offs available within the problem.

Provided the objective functions of the Hamiltonian are normalised, the relative magnitudes of the coupling constants roughly encode the relative importance of the objectives.
This is another reason why we use linear objective functions, since their curvature then does not influence the interpretation of coupling constants as proxies for relative importance.
However, this interpretation is not perfect - certain objectives may need less weight to satisfy, and using the entire theoretical range of an objective function rather than the range it might feasibly visit during optimisation distorts it further. This is further complicated by correlations between objective functions; consider the relationship between compactness and county boundaries. Satisfying county boundaries also contributes to ensuring compactness, owing to the fact that counties do not have the salamandering shape that compactness seeks to avoid.

It should also be noted that multiplying every
coupling constant and the temperature by a common positive factor
$\kappa$ leaves the MCMC dynamics unchanged, since
$\Delta H / T = (\kappa\Delta H)/(\kappa T)$ for every proposed move.
This $\kappa$-symmetry means that only the ratios of coupling constants and temperature to one reference coupling constant $J_\alpha$ are meaningful. We exploit this by fixing $J_P = 1$ as a
reference scale and treating $J_C$, $J_D$, and
$J_B$ as the free external parameters. This choice is a labelling convention, not a normative claim: setting $J_P=1$ does not privilege $H_P$ in any way that a re-scaling of the other constants could not also achieve.

In a future work, we will examine the use of particular distance measures in determining the ``optimal'' configuration on the Pareto front. However, in the absence of a legally affirmed basis for the linear form of our Hamiltonians, we expect the Pareto front to change shape if we parametrise our objective functions differently, and so pointing to an optimal configuration at this stage would be premature. Currently, we focus on determining the Pareto front, so we achieve the set of configurations which minimally compromise on the objective functions, but which are not definitively better than each other. 

\section{Results}
\subsection{Potts Model Convergence}
\begin{figure}
    \centering
    \includegraphics[page=1,width=\linewidth]{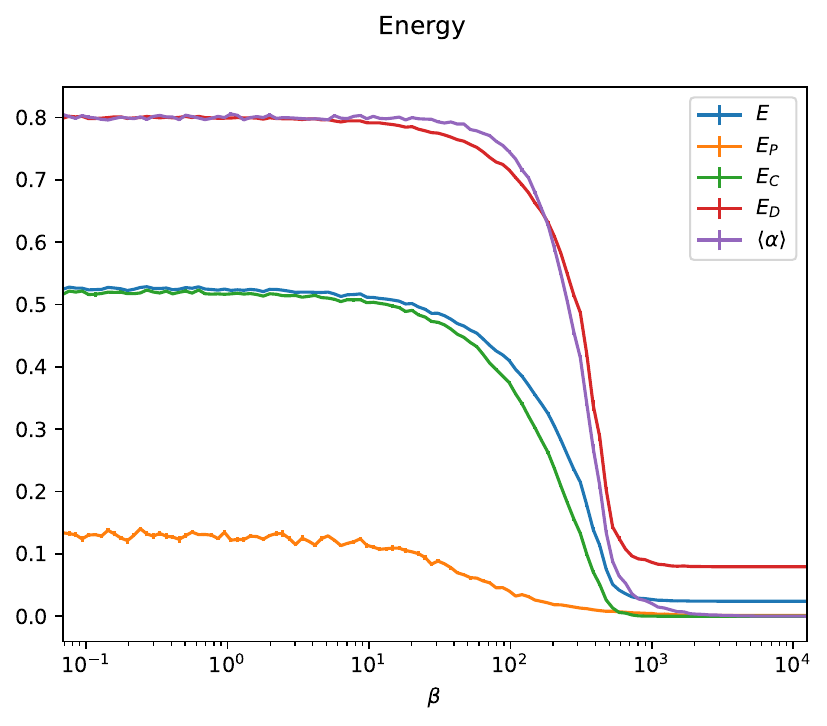}
    \caption{Plotted are the values for the individual terms in a typical COTHROM Hamiltonian as a function of inverse temperature $\beta$. The $\langle \alpha\rangle$ legend item is the acceptance rate of the MCMC algorithm following a proposed ED change, which decreases to $0$ at low temperature.}
    \label{fig:SimulatedAnnealling}
\end{figure}

\begin{figure}
    \centering
    \includegraphics[page=1,width=\linewidth]{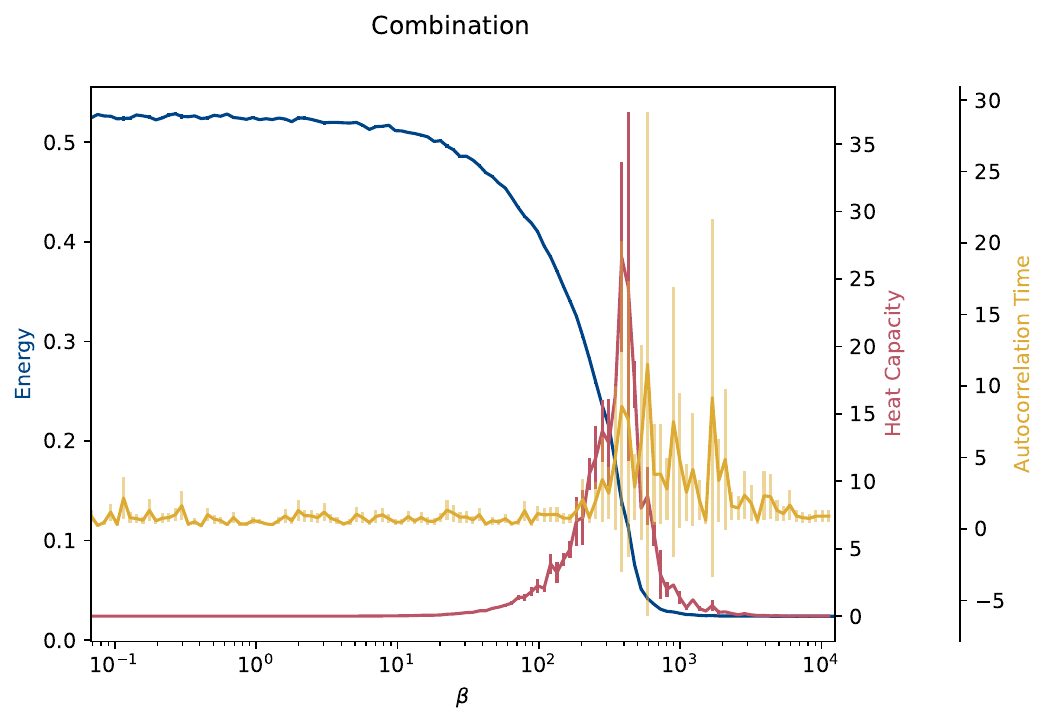}
    \caption{A plot capturing the convergence behaviour of the algorithm as a function of temperature. As the system cools, the system energy decreases, and as the system transitions from unstructured to structured constituencies, there is a spike in autocorrelation time (a measure of the independence of MCMC sample - larger roughly equates to greater sampler dependence and worse sampling efficiency), which is characteristic of the phase transitions present in the physics model on which COTHROM relies. Heat capacity is a diagnostic borrowed from statistical physics which measures the fluctuation of the log-probability of the target distribution across the chain's samples.}
    \label{fig:Convergence}
\end{figure}

In \autoref{fig:SimulatedAnnealling}, we observe the convergence of each of the terms in our Hamiltonian, evident by the energy plateaus. Assessing the configuration visually for different temperatures, and observing the transition from random to structured solutions, indicates that the system is undergoing the equivalent of a phase transition in statistical physics, whereby constituencies ``crystallise'' out of random, unbiased initial configurations following annealing. \autoref{fig:Convergence} shows how the sampling efficiency changes as the simulation proceeds, in line with what is expected for such a lattice system. 

\subsection{MCDA Results}
\begin{figure}
    \centering
    \includegraphics[page=1,width=\linewidth]{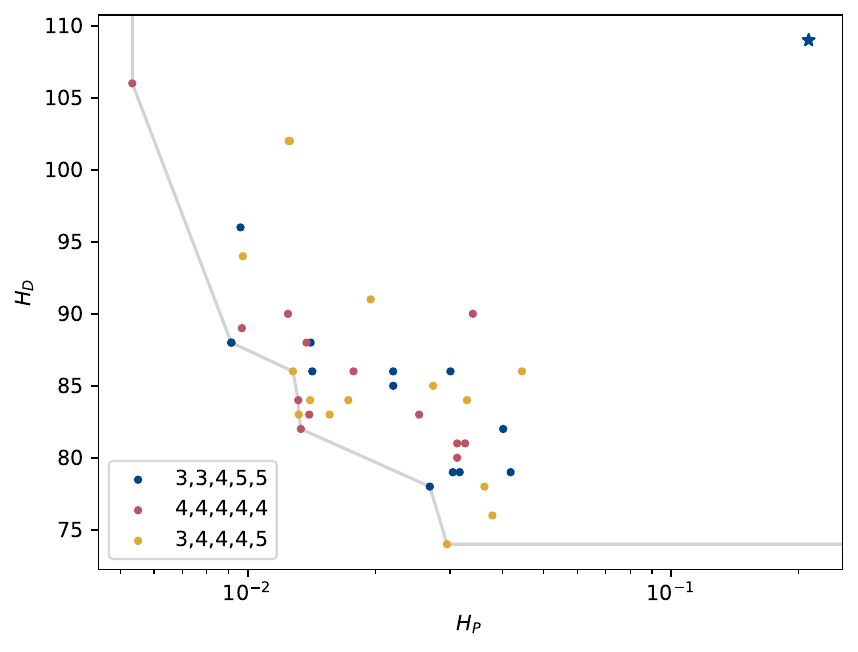}
    \caption{The Pareto front (denoted by the grey line) arising from considering only proportional representation and compactness for county Cork, for multiple seat configuration options and objective weightings. The star in the top right is the 2023 legal boundaries' performance on the $H_P$ and $H_D$ metrics. As noted in the text, the EC cater to other constraints not considered here, which partly explain the significant departure for Pareto optimality.}
    \label{fig:Pareto}
\end{figure}

\begin{figure}
    \centering
    \includegraphics[page=1,width=\linewidth]{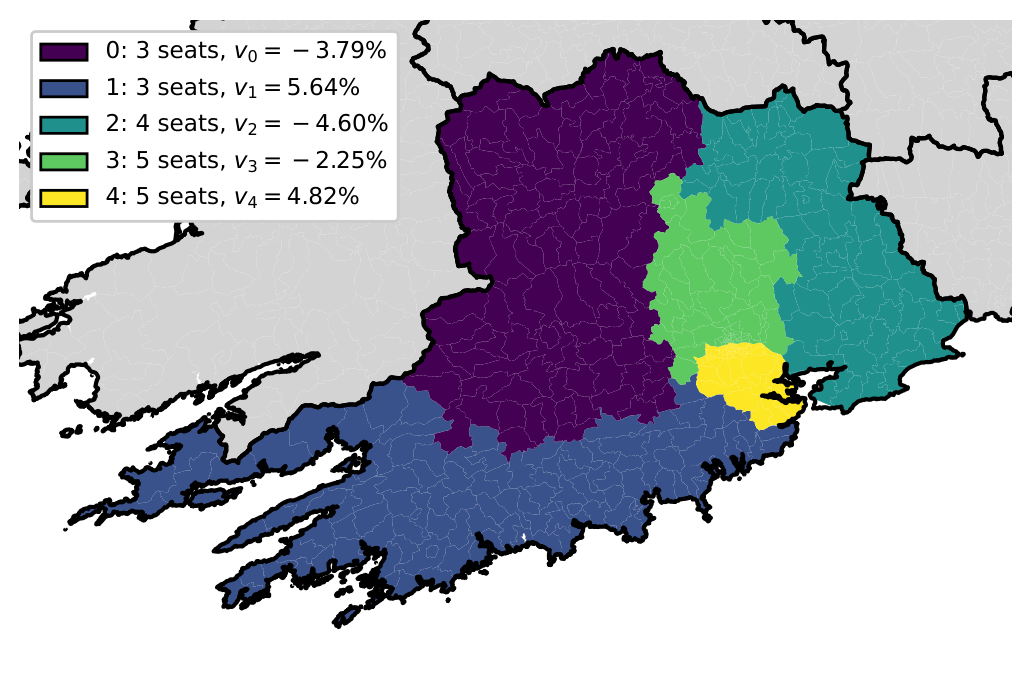}
    \caption{The 2023 electoral boundaries for Cork. The values in the legend indicate the departure from proportional representation for each constituency.}
    \label{fig:ECCORK}
\end{figure}

\begin{figure}
    \centering
    \includegraphics[page=1,width=\linewidth]{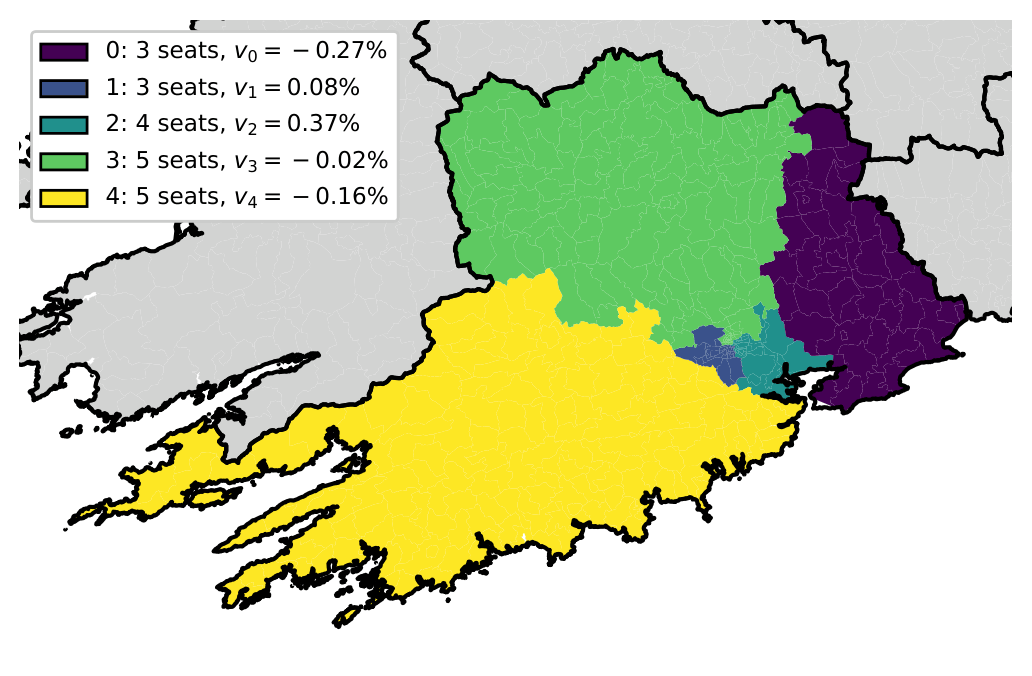}
    \caption{The electoral boundaries generated by COTHROM, using the objective weightings corresponding to the leftmost $\{3,3,4,5,5\}$ point in the Pareto plot in Fig. \ref{fig:Pareto}. The values in the legend indicate the departures from proportional representation for each constituency, which are significantly lower than that of the EC's. However, no attempt was made to bias the search towards the current boundaries in favour of continuity over optimality, although this may be achieved by annealing from the current configuration and a lower temperature. No was any consideration given to city limits/mountain ranges/rivers, although this may also be encoded in a future work.}
    \label{fig:COTHROMCORK}
\end{figure}

We focus on County Cork as a proof of concept of the MCDA methodology, since performing a full Pareto front analysis for the entire 3,420 ED system and all objectives is beyond the computational resources we can currently deploy. We consider five 3-, 4-, and 5-seater constituencies with 20 total seats, based on the current two 3-seater (Cork North-West, Cork South-West), one 4-seater (Cork East), and two 5-seater (Cork North-Central, Cork South-Central) constituencies within Cork.

By considering constituencies entirely within a county, we remove the need to optimise the corresponding county boundary Hamiltonian, i.e.\ $H_B = 0$. Furthermore, as contiguity is a strict requirement for constituencies, we discard configurations with non-contiguous constituencies and only consider cases where $H_C=0$; we achieve this by setting the contiguity coupling equal to half of the sum of coupling constants. We consider a range of values between 0 and 2.5 for the compactness coupling constant, and fix the population coupling at unity as previously discussed. The corresponding set of coupling constants we consider is then given by
\begin{align}
    \label{eq:MCDA_couplings}J_P &= 1, & J_C &= 1 + J_D, & J_D &\in [0, 2.5], & J_B &= 0.
\end{align}
For each of these sets of couplings we execute our simulated annealing algorithm using the Gibbs sampler with a hot (uniformly random) initial configuration and 150 sweeps between temperatures. We repeat this for all possible combinations of seat allocations for 20 seats across five constituencies, and identify configurations which compose the Pareto front in the $H_DH_P$-plane.

Our Pareto front plot, along with a point for the 2023 Cork constituencies, is shown in \autoref{fig:Pareto}. \autoref{fig:ECCORK} and \autoref{fig:COTHROMCORK} show the 2023 Cork constituency boundaries and a Pareto-optimal configuration found with COTHROM, respectively, along with the population variance for each constituency in both cases. We remind the reader once again that while these results represent an approximate algorithmic optimum for a given choice of objective weightings, the results are not subject to all of the constraints the EC contend with. There are implicit objectives that the EC address stemming from the public consultations, communities not wanting to be split, and so on. We also do not consider geographical features like mountains and rivers. 
In a future work the geographical and community connections could be encoded in a manner similar to the county boundaries term. Such a community boundaries term would enable the EC to rigorously and transparently encode insight from the public consultation. However, it is not an explicit objective in the Irish constitution or the Electoral Boundaries Act, and it would require its own coupling constant/weight compared to the county boundaries given its lack of legal standing. Acting on a community boundaries term also risks algorithmic endogeneity - the idea that the governance decisions can shape the inputs to the algorithm going forward. If a community is consistently used in a county breach to fulfil proportional representation, that community may in time feel an affinity for being represented in a county that is not theirs.


\subsection{County Boundaries}
\begin{figure}
    \centering
    \includegraphics[width=0.49\linewidth]{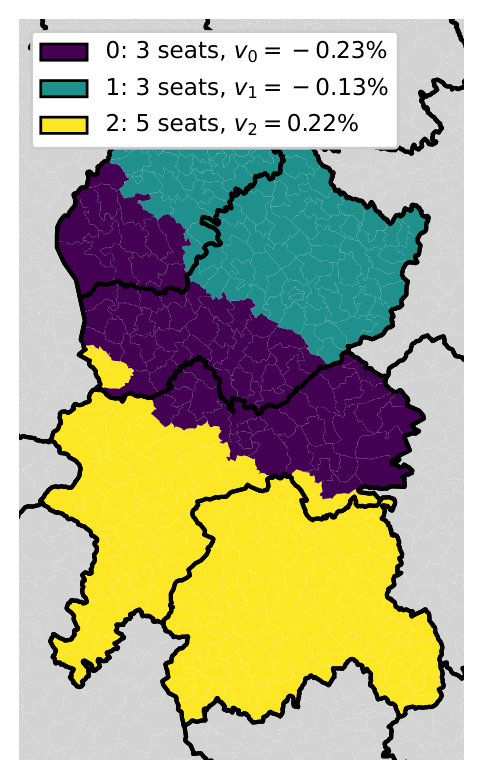}
    \includegraphics[width=0.49\linewidth]{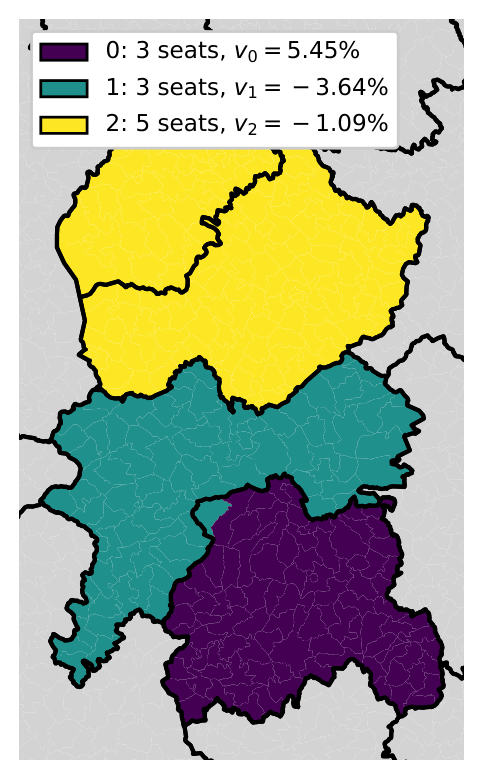}
    \caption{Electoral boundaries generated by COTHROM for midland counties with two 3-seater and one 5-seater constituencies. Left (Right): optimal configuration for $J_B=0$ ($J_B=2.5$). Note that for the left configuration, the single Westmeath ED in the 5-seater constituency is considered to be connected at a single point to the rest of the constituency.}
    \label{fig:goodbad_county}
\end{figure}
As discussed in \autoref{App:CountyBreaches}, our simple definition of a county breach in this work is not reflective of a ``true'' county boundary breach, as a quantitative definition is unclear. We nonetheless demonstrate that our framework can incorporate county boundaries in some regard by analysing results for the midland counties (Longford, Westmeath, Offaly, Laois). \autoref{fig:goodbad_county} shows two optimal configurations at zero and large county boundary coupling constant $J_B$; the former has excellent proportional representation but no regard for county boundaries, and the latter matches boundaries nearly exactly at the expense of a much higher population variance. Note that, although the configuration for large county boundary coupling nearly exactly reproduces the 2023 boundaries for the Longford-Westmeath, Offaly, and Laois constituencies, the entirety of County Longford is considered as breached in the single 5-seater constituency (right side of \autoref{fig:goodbad_county}, yellow area) due to our underdeveloped county boundary term.
\subsection{Policy Implications}\label{sec:eval-policy}
Several major policy implications emerge:
\begin{enumerate}
    \item \textbf{Optimality \& Efficiency of redistricting}: When this approach is scaled to the national level, and physical features such as mountains and rivers are encoded, this will enable the rapid generation of mathematically optimised configurations. These may serve as a starting point to then begin incorporating concerns raised during the public submission process.
    \item \textbf{Guaranteed geographically equitable treatment}: The same objective function acts across the EDs under consideration, so that, for example, once the relative importance of proportional representation is decided, it is enacted with equal emphasis in across the geography considered.
    \item \textbf{Historical information}: Currently, the algorithm initiates from a blank slate - a completely random configuration of constituencies, in the interest of performing a thorough optimisation. In practice, temporal continuity will feature in redistricting consideration. Fortunately, our approach will enable the enable us to bias towards the current constituency arrangement. We may do this by initialising the annealing process from a lower temperature, and starting with the existing configuration.
    \item \textbf{Analyse trade-offs of objectives}: The statutory language (i.e., "as far as practicable") leaves the relative weighting of objectives undefined. Our framework operationalises this by making the trade-offs explicit: the Pareto front quantifies how much county-boundary integrity must be sacrificed for a given gain in proportional representation, and vice versa. This means a proposed configuration can be checked for dominance, with any plan that is strictly dominated indicating that an admissible alternative exists which is at least as good on every objective and strictly better on one.
    \item \textbf{Reproducibility \& auditability}: All optimisation parameters can be published alongside the proposed configurations, and the code itself open-sourced. This contrasts with the current paradigm, where the EC's judgement that a configuration is compliant with the terms of reference is not accompanied by any demonstration that no admissible alternative satisfies those terms more effectively (since the means do not exist to perform such a search currently). Independent researchers, the courts, or members of the public could reproduce the optimisation and inspect each interpretive choice that fed into it.
\end{enumerate}

There are also many future practical considerations we will endeavour to tackle with this framework:
\begin{enumerate}
    \item \textbf{Future-proofing constituencies}: There is scope to combine our framework with a geo-spatially resolved population dynamics model to generate constituencies that are maximally stable across the twelve-year revision window. This would partially mitigate the temporal instability that the EC is already expected to manage, and could reduce the disruption associated with each revision cycle. Any such forecasting introduces its own assumptions about migration and demographic change, which would need to be made transparent in the same manner as the constitutional objectives.
    \item \textbf{Varying total seat number}: The Electoral Reform Act 2022 permits between 171 and 181 representatives. We aim for our framework to sweep this range systematically and characterise the trade-offs at each seat number.
    \item \textbf{Varying constituency magnitude}: This framework can be trivially extended to consider six-seat constituencies. This will position the framework as a tool for evidence-based input into any future legislative review of the magnitude rule.
    \item \textbf{Encoding deliberative input}: Concerns raised through the public consultation process (i.e. community ties, geographical features, the desire of particular regions to remain together) could in principle be encoded as additional terms in the Hamiltonian with their own coupling constants, ensuring that they are applied transparently and consistently across the country rather than negotiated case-by-case. This would, however, require legal grounding that does not currently exist, and carries a risk of algorithmic endogeneity: governance decisions taken today can shape the community-affinity inputs the algorithm sees tomorrow.
    \item \textbf{MCDA implications for legally equally stated objectives}: The Constitution and the Electoral Reform Act state the soft objectives in formally equal terms, but MCDA, once the functional form of the objective functions are made certain, will point to substantive optimisation requiring the coupling constants to be unequal. The current redistricting paradigm in Ireland has no means to articulate this tension, since trade-offs are made implicitly in deliberation rather than surfaced explicitly. To the authors' knowledge, the question of whether an unequal weighting of formally equal legal objectives is itself constitutionally permissible has not been addressed in the Irish legal context.
    \item \textbf{Legal defensibility}: If a constituency revision is challenged, the EC is not presently required to demonstrate that the chosen plan is near-optimal under the terms of reference. The framework provides a route to that demonstration: the EC could evidence that the published plan lies on (or near) the Pareto front under a stated weighting, and that any alternative configuration improving one objective necessarily worsens another. More broadly, because competing redistricting objectives are embedded within the same constrained mathematical system, an open methodological question remains: when multiple constituency configurations satisfy the legal constraints, is there any principled basis for preferring one valid configuration over another?
\end{enumerate}

\section{Conclusion}
The rules which direct the Irish redistricting process implicitly give form to a mathematical landscape which, until now, has not been utilised to improve the redistricting process. In this work, we have demonstrated the first application of the Potts model to Irish redistricting and a PR-STV system, encompassing proportional representation, compactness, contiguity, and county boundary objectives. 

Analysing the structure of this optimisation landscape in an optimisation setting provides insight in into how these rules interact, and where tensions between democratic objectives arise. We developed an MCDA framework for analysing objective trade-offs, which, when deployed in Ireland's county Cork as an example, reveals a Pareto front structure illustrating the extent of departure of Ireland's current electoral boundaries from optimality. In a future work, this will be scaled to a full-Republic analysis, including the county boundaries objective as an additional axis over which to construct the Pareto front. Physical features, such as mountains and rivers, and community input via public submissions to Ireland's Electoral Commission are also left to a later work.

In terms of democratic integrity, COTHROM bolsters the legitimacy of the boundaries it produces by virtue of its reproducibility, auditability, transparent encoding of objectives, equal enforcement of legal objectives across regions, and the use of reliable MCMC methods.

\pagebreak
\bibliographystyle{unsrtnat}
\bibliography{Bibliography}

\appendix
\newpage
\appendix
\section{Breaches of County Boundaries}\label{App:CountyBreaches}
\begin{figure}[htp]
\centering
\includegraphics[width=\textwidth]{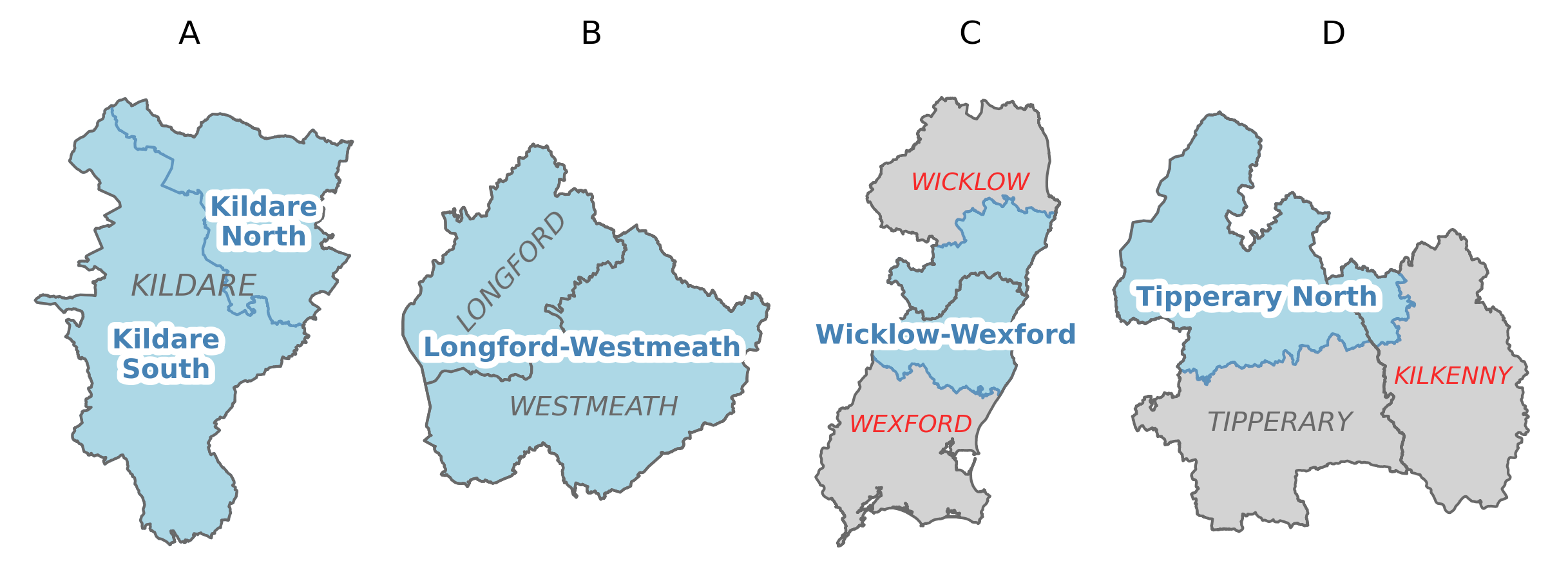}
\caption{Examples of constituency configurations, illustrating situations where county boundaries are, or are not, considered to be breached by the Electoral Commission. Constituencies are shown in blue and counties are shown in grey, with breached counties labelled in red.}
\label{fig:example-configurations}
\end{figure}

Section 57.2 (c) of the Electoral Reform Act 2022 requires that
\begin{quote}
    ``the breaching of county boundaries shall be avoided as far as practicable''
\end{quote}
during reviews of constituency boundaries \cite{ElectoralReformFullAct2022}. However, in the absence of a precise definition of what constitutes a county boundary breach, it is not always clear whether a given configuration of constituencies should be considered to breach a particular county boundary. For example, Table \ref{tab:county-boundary-breaches} shows the county boundaries which the Electoral Commission considered to be breached in each of the last four constituency revisions, as set out in Table 4 of their Constituency Review Report 2023 \cite{AnCoimisiunToghchain2023}. Focusing on the 2023 revision, we note that the Electoral Commission lists six counties whose boundaries they consider to be breached, namely Donegal, Galway, Kilkenny, Meath, Wexford, and Wicklow \cite[Table~4]{AnCoimisiunToghchain2023}. By comparing this classification to the precise configuration of the 2023 constituencies, we can reasonably infer that a constituency is not considered to breach county boundaries if it is wholly contained within a county (example A in Figure \ref{fig:example-configurations}), or if it comprises the entirety of two counties (example B in Figure \ref{fig:example-configurations}). However, not all constituency configurations are so simple, and this criterion is not sufficient to categorise all counties as either ``breached'' or ``unbreached''. For example, we might assume that Wicklow and Wexford are both considered to be breached because the Wicklow-Wexford constituency overlaps with both of these counties without wholly containing either (example C in Figure \ref{fig:example-configurations}). However, it is unclear why the same does not apply to, say, the Tipperary North constituency, which overlaps with both Kilkenny and Tipperary without wholly containing either -- in this case, the EC only considers the boundary of Kilkenny to be breached (example D in \ref{fig:example-configurations}).

\begin{table}[ht]
\centering
\caption{Breaches of county boundaries occurring in the last four constituency boundary revisions, according to the Electoral Commission's Constituency Review Report 2023 \cite{AnCoimisiunToghchain2023}.}
\label{tab:county-boundary-breaches}
\begin{tabular}{p{1.5cm}p{1.8cm}p{7cm}}
\toprule
\textbf{Year of \newline Revision} & \textbf{Number of \newline Breaches} & \textbf{Counties whose boundaries were breached} \\
\midrule
2023 & 6 & Donegal, Galway, Kilkenny, \newline Meath, Wexford, Wicklow \\[1em]
2017 & 10 & Donegal, Galway, Laois, Mayo, Meath (2), \newline Offaly, Roscommon, Tipperary, Westmeath \\[1em]
2013 & 10 & Carlow, Cavan, Clare, Donegal, Galway,\newline Kildare, Mayo, Meath, Tipperary, Westmeath \\[1em]
2009 & 9  & Carlow, Clare, South Tipperary, Waterford,\newline Leitrim, Westmeath, Limerick, Meath, Offaly \\
\bottomrule
\end{tabular}
\end{table}

To deal with these issues of definition, it would be useful to devise a rigorous, quantitative method of determining whether or not a particular configuration of constituencies breaches a particular county boundary. Such a method should be able to reliably classify all counties as ``breached'' or ``unbreached'' by a specific constituency revision in agreement with the classification made by the Electoral Commission.

\begin{figure}[htp]
\centering

\includegraphics[height=0.88\textheight, keepaspectratio]{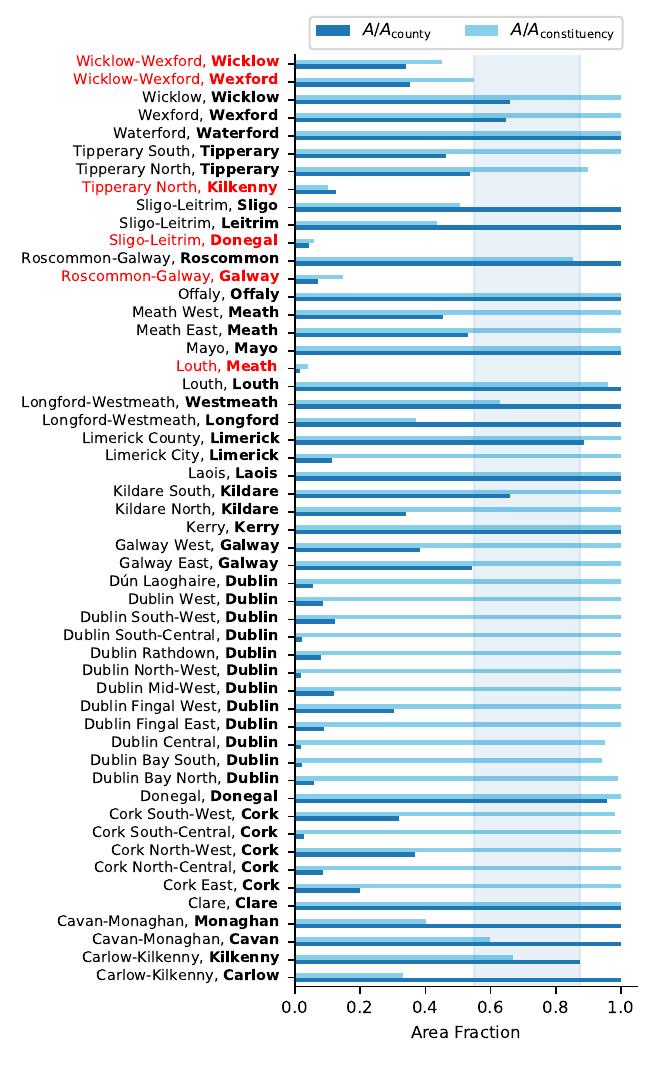}
\caption{The area of each constituency-county overlap as a fraction of the total constituency or county area for the 2023 constituency boundaries. Constituency names are in regular typeface, and county names are in bold. Overlaps labelled in red correspond to county boundary breaches.}
\label{fig:bar-2023}
\end{figure}

\begin{figure}[htp]
\centering
\includegraphics[height=0.88\textheight, keepaspectratio]{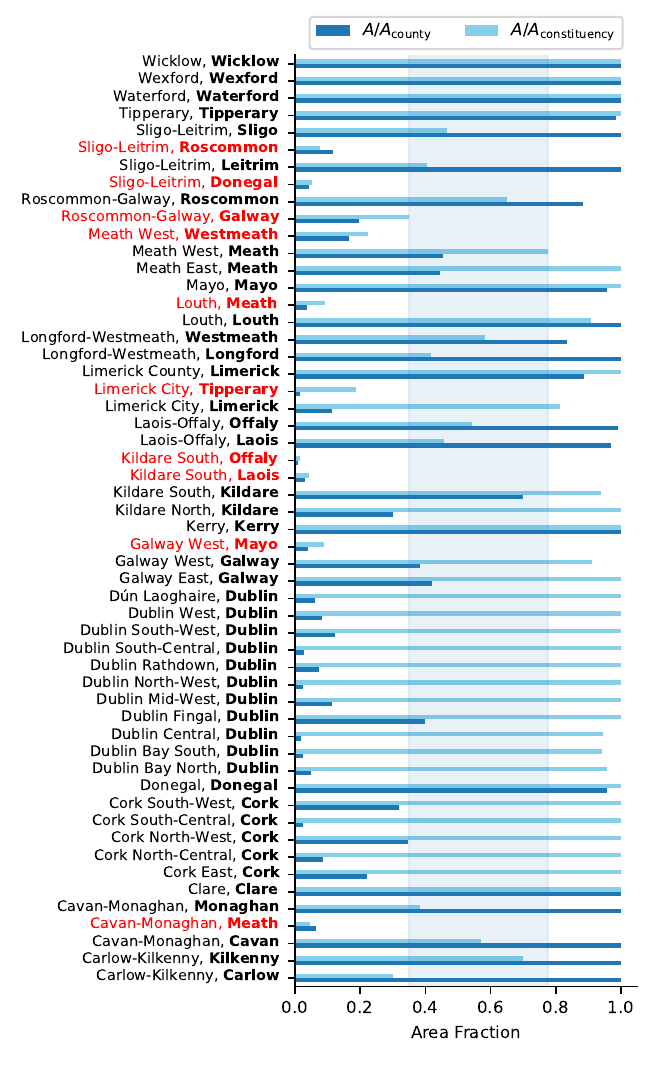}
\caption{The area of each constituency-county overlap as a fraction of the total constituency or county area for the 2017 constituency boundaries. Constituency names are in regular typeface, and county names are in bold. Overlaps labelled in red correspond to county boundary breaches.}
\label{fig:bar-2017}
\end{figure}

\begin{figure}[htp]
\centering
\includegraphics[height=0.88\textheight, keepaspectratio]{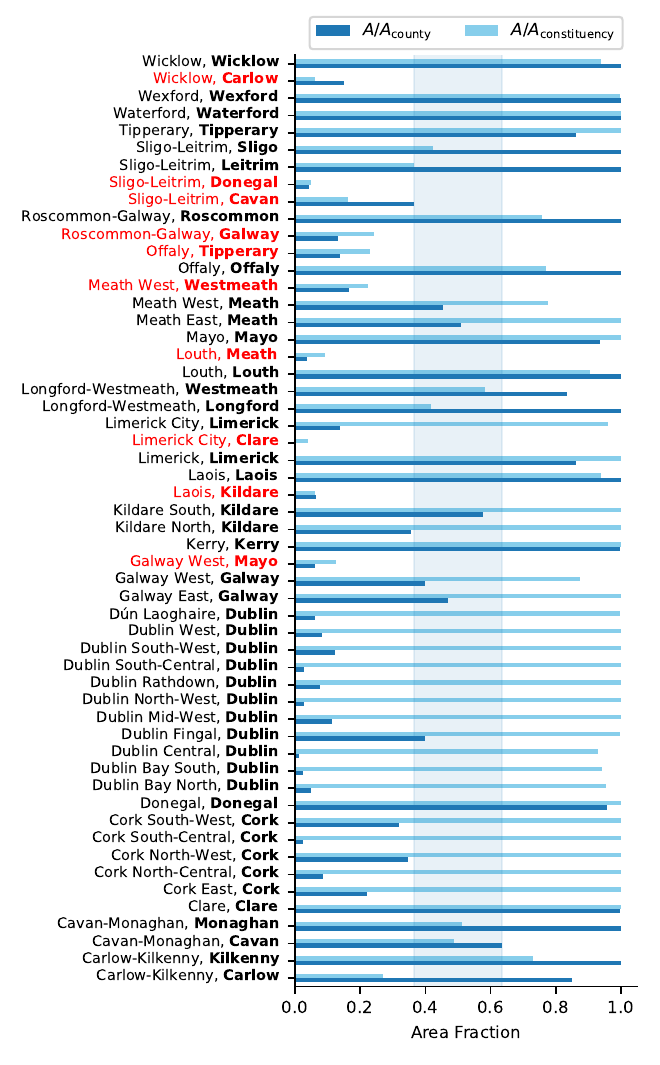}
\caption{The area of each constituency-county overlap as a fraction of the total constituency or county area for the 2013 constituency boundaries. Constituency names are in regular typeface, and county names are in bold. Overlaps labelled in red correspond to county boundary breaches.}
\label{fig:bar-2013}
\end{figure}

\begin{figure}[htp]
\centering
\includegraphics[height=0.88\textheight, keepaspectratio]{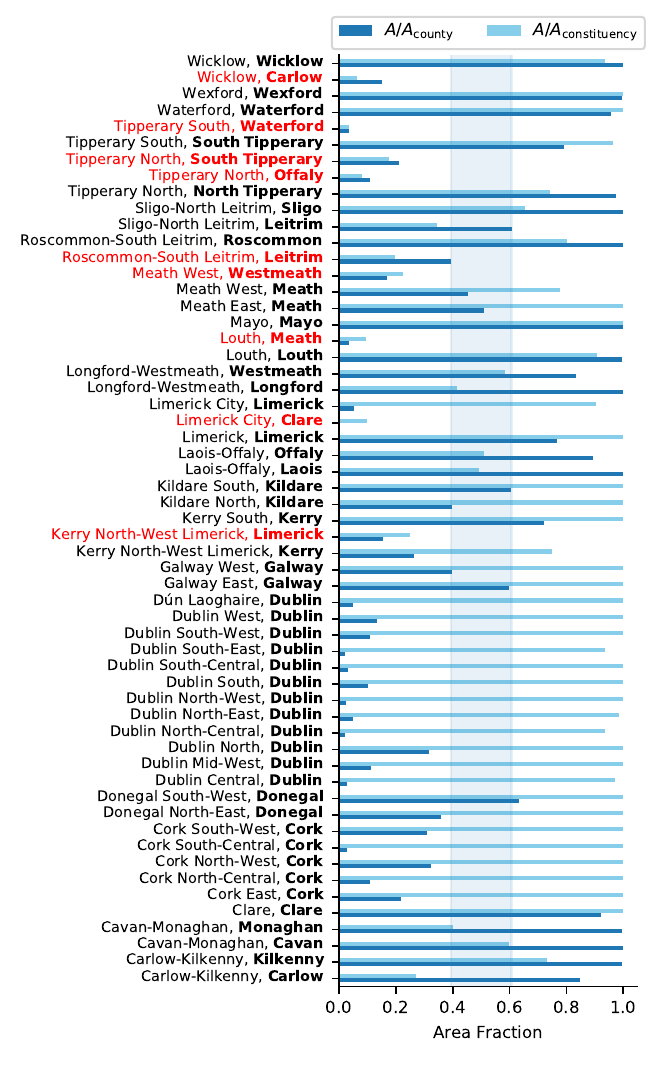}
\caption{The area of each constituency-county overlap as a fraction of the total constituency or county area for the 2009 constituency boundaries. Constituency names are in regular typeface, and county names are in bold. Overlaps labelled in red correspond to county boundary breaches.}
\label{fig:bar-2009}
\end{figure}

In the section of their Constituency Review Report 2023 \cite{AnCoimisiunToghchain2023} that discusses their approach to the avoidance of county boundary breaches, the Electoral Commission notes that
\begin{quote}
    ``The transfer of a small population can add to the perception of lack of effective representation. A view was expressed in submissions that if a county boundary was to be breached, the breach should involve the transfer of a sufficiently large number of persons to ensure that the transferred population carried some weight." 
\end{quote}
This suggests that any redistricting process conducted by the Electoral Commission should minimise the number of voters who are assigned to a constituency in which they form a small minority from their own county -- that is, that a county boundary should not be considered truly breached by a constituency if the apparently-breached region forms a large enough proportion of the constituency as a whole, so that voters in this region are not isolated from the rest of their county. This idea can be captured quantitatively by performing an overlay of all constituency boundaries with all county boundaries, in order to divide the whole country into overlapping constituency-county pairs. For each such constituency-county pair, one may then compute the area of the overlapping region as a proportion of the entire constituency's area, and as a proportion of the entire county's area. Carrying out this procedure for the last four constituency revisions, in 2023, 2017, 2013, and 2009, we obtain the results shown in Figures \ref{fig:bar-2023}, \ref{fig:bar-2017}, \ref{fig:bar-2013}, and \ref{fig:bar-2009}.

Inspecting each of these figures for features which may enable counties to be classified as breached or unbreached, we note that there is, in fact, a clear method of splitting constituency-county pairs into two distinct groups. In particular, if we separate out the constituency-county pairs for which both \(A/A_{\mathrm{constituency}}\) \emph{and} \(A/A_{\mathrm{county}}\) are sufficiently low, then the counties involved are exactly those labelled as breached by the Electoral Commission (these are labelled in red in the plots). The precise definition of what it means for an area fraction to be sufficiently low can be determined by a choice of threshold for both \(A/A_{\mathrm{constituency}}\) and \(A/A_{\mathrm{county}}\), where we label all constituency-county pairs having \emph{both} quantities below their respective thresholds as breached. This leads to the rule:
\begin{defn}[Tentative]
A constituency \(C\) breaches the boundary of county \(K\) if the area of overlap \(C\cap K\) as a proportion of the total area of constituency \(C\) is less than \(T_{\mathrm{constituency}}\) \emph{and} the area of overlap \(C\cap K\) as a proportion of the total area of county \(K\) is less than \(T_{\mathrm{county}}\).
\end{defn}
For example, in 2023, the maximum possible \emph{simultaneous} ranges of the thresholds are \(0.55\leq T_{\mathrm{constituency}}, T_{\mathrm{county}} \leq 0.87\) -- any lower and we do not always get all six counties listed as breached by the Electoral Commission; any higher and we may get extraneous counties beyond those listed (although note that other threshold pairs may be possible if we go beyond the range of one of the two thresholds but not the other, as discussed below). Similar thresholds can be derived for the other years; these threshold regions are shaded in Figures \ref{fig:bar-2023}, \ref{fig:bar-2017}, \ref{fig:bar-2013}, and \ref{fig:bar-2009}. In particular, note that this metric accurately captures the fact that the boundaries of Meath are breached twice by the 2017 constituency revision: once by Louth, and once by Cavan-Monaghan.

\begin{figure}[htp]
\centering
\begin{subfigure}[t]{0.49\textwidth}
\includegraphics[width=\textwidth]{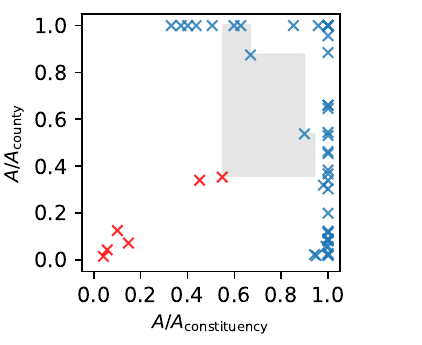}
\caption{2023 Boundaries}
\end{subfigure}
\begin{subfigure}[t]{0.49\textwidth}
\includegraphics[width=\textwidth]{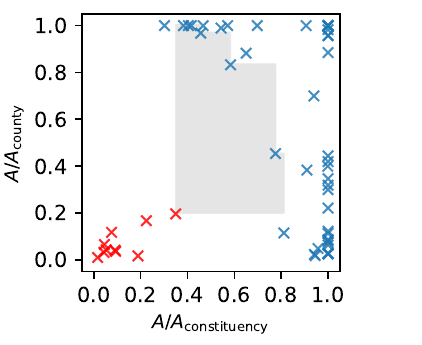}
\caption{2017 Boundaries}
\end{subfigure}

\begin{subfigure}[t]{0.49\textwidth}
\includegraphics[width=\textwidth]{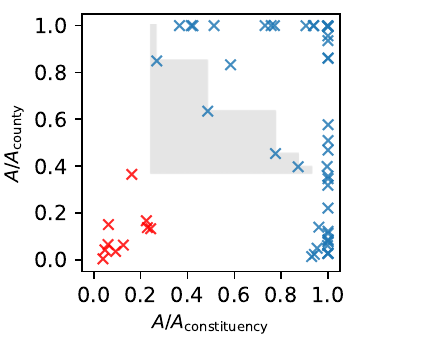}
\caption{2013 Boundaries}
\end{subfigure}
\begin{subfigure}[t]{0.49\textwidth}
\includegraphics[width=\textwidth]{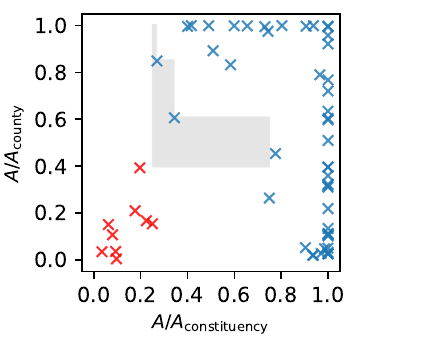}
\caption{2009 Boundaries}
\end{subfigure}
\caption{Area fractions of overlapping constituency-county regions for 2023, 2017, 2013, and 2009. The overlap regions corresponding to county breaches are shown in red. The shaded regions represent the possible thresholds \((T_{\mathrm{constituency}}, T_{\mathrm{county}})\) which can be chosen so that the pairs with \(A/A_{\mathrm{constituency}} \leq T_{\mathrm{constituency}}\) and \(A/A_{\mathrm{county}} \leq T_{\mathrm{county}}\) correspond exactly to the breached counties for that year as classified by the Electoral Commission.}
\label{fig:scatter-plots}
\end{figure}

This data can be more intuitively represented using a two dimensional plot rather than a bar chart, as shown in Figure \ref{fig:scatter-plots}.

\begin{figure}[htp]
\centering
\includegraphics{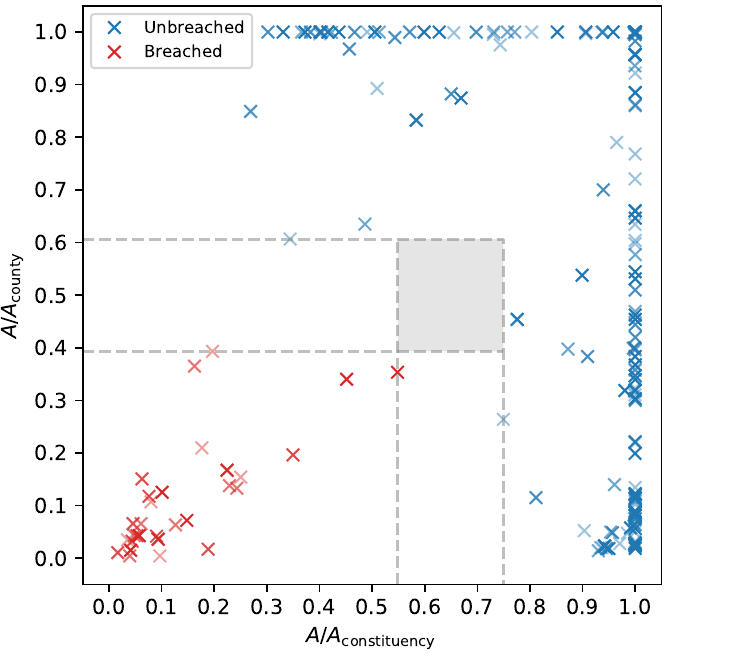}
\caption{Area fractions of overlapping constituency-county regions for 2023, 2017, 2013, and 2009 (darker \(=\) more recent). The shaded region represents the common thresholds \((T_{\mathrm{constituency}}, T_{\mathrm{county}})\) which can be chosen for all years so that the pairs with \(A/A_{\mathrm{constituency}} \leq T_{\mathrm{constituency}}\) and \(A/A_{\mathrm{county}} \leq T_{\mathrm{county}}\) correspond exactly to the breached counties.}
\label{fig:scatter-plot-all-years}
\end{figure}

One may ask whether there is any range of possible thresholds which is consistent across all four available years, and indeed by overlaying the four separate threshold grids we find that there is a consistent threshold range of \(0.5483 \leq T_{\mathrm{constituency}} \leq 0.7494\) and \(0.3932 \leq T_{\mathrm{county}} \leq 0.6065\). This region is shown in Figure \ref{fig:scatter-plot-all-years}. This means that our tentative quantitative definition of a county boundary breach can be applied consistently across all available years for these select threshold values.

It should be remarked that any definition implementing a binary classification of counties as either breached or unbreached is necessarily going to be slightly artificial. While this approach based on area fractions effectively captures the extent to which a voting community is isolated from the rest of their county, there is less evidence to support the choice of a particular hard cutoff that splits the counties in such a definite way. We have focused here on determining the thresholds that reproduce the exact binary classification provided by the Electoral Commission, but these hard cutoffs are necessarily ad-hoc. It may be more natural to consider the area fraction pairs as providing a continuous measure of the extent to which a county boundary is breached by a particular configuration of constituencies, with configurations where more constituency-county pairs occur closer to the bottom left of a plot of \(A/A_{\mathrm{county}}\) versus \(A/A_{\mathrm{constituency}}\) being penalised proportionally more by the redistricting algorithm. 

Finally, we note that our approach here is based on the area of overlapping constituency-county regions. At the scale we are considering, area is a good proxy for population, and very similar results are obtained if the constituency, county, and overlapping-region areas are replaced by their respective population counts (which can be easily derived from the underlying electoral division populations), yielding the same overall classification with only a slight adjustment of the precise threshold values. Nevertheless, using population fractions instead of area fractions might be more conceptually appealing as it more accurately reflects the social impact of county boundary breaches on individual voters.

\section{Are Electoral Divisions Indivisible?}\label{Appendix:EDBreaking}

\begin{figure}[H]
    \centering

    \begin{subfigure}{\linewidth}
        \centering
        \includegraphics[width=0.7\linewidth]{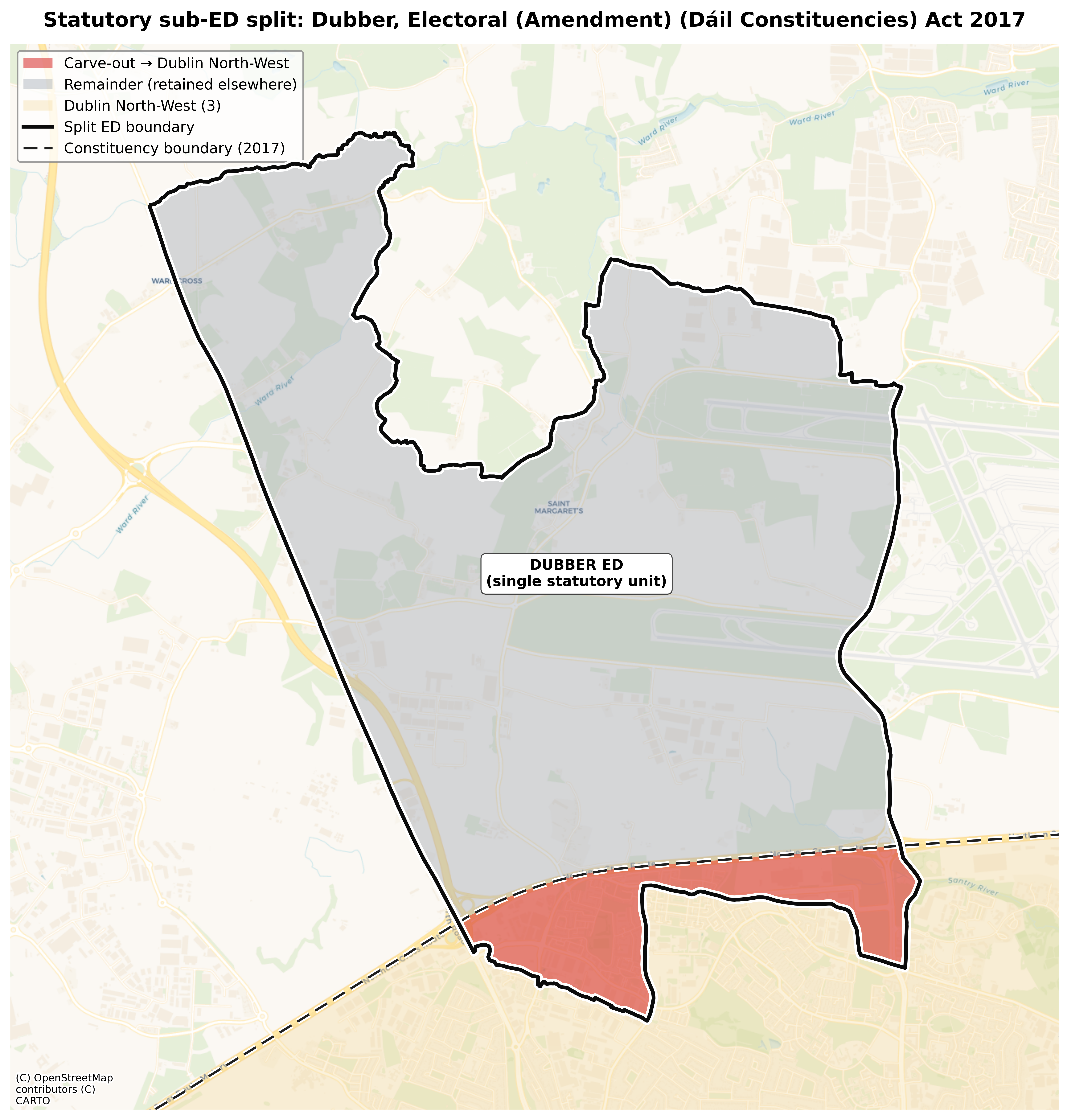}
        \caption{Dubber ED, 2017 Act: Under the Electoral (Amendment) (Dáil Constituencies) Act 2017, the Dubber Electoral Division is bisected along the M50 Northern Cross Route by the constituency boundary (dashed); the southern fragment (red) is assigned to Dublin North-West (sand) and the northern remainder (grey) is retained by a neighbouring constituency in Fingal.}
        \label{fig:dubber1}
    \end{subfigure}

    \vspace{0.5cm}

    \begin{subfigure}{\linewidth}
        \centering
        \includegraphics[width=0.9\linewidth]{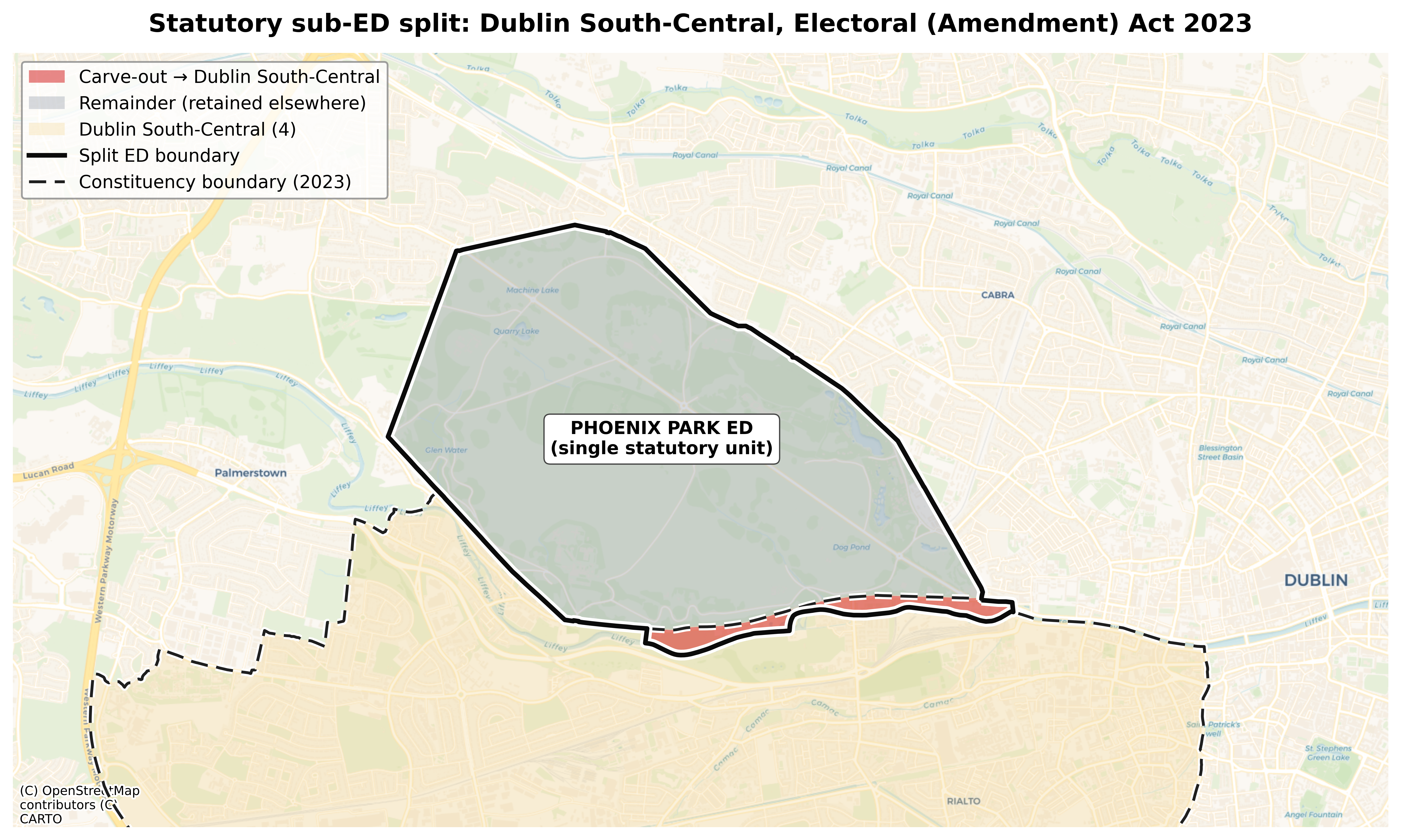}
        \caption{Phoenix Park ED, 2023 Act: Under the Electoral (Amendment) Act 2023, the Phoenix Park Electoral Division is split along its southern edge, with the carved fragment (red) assigned to Dublin South-Central (sand) and the remainder (grey) retained elsewhere. }
        \label{fig:dubber2}
    \end{subfigure}

    \caption{Statutory sub-ED splits illustrating weakness of assumption of ED indivisibility. In both cases the cut follows a named physical feature rather than any ED boundary. }
    \label{fig:dubber_combined}
\end{figure}

Our graph representation rests on the assumption that the Electoral Division is the indivisible atomic unit of the partition: every move in the Markov chain reassigns a whole ED, so if the admissible space permits sub-ED splits, the configuration space we explore is a strict subset of the statistical ones. Two distinct notions of an ED are in circulation. The legally defined EDs (3,440 in number) are the units recognised in primary legislation and statutory instruments. The statistical EDs (3,420) are adaptations published by the Central Statistics Office for reporting and disclosure control, and the set is not stable across census releases. The Electoral Commission's 2023 Constituency Review Report states ~\cite{AnCoimisiunToghchain2023} that the work began ``with consideration of the maps of the current constituencies, and the Census figures for every Electoral Division,'' and elsewhere~\cite{AnCoimisiunToghchain2023} notes figures ``omitted due to disclosure control practices in line with CSO data-use'' phrasing that applies to the statistical, not the legal, set. Which set the Commission treats as definitive (and actually uses) is therefore not stated unambiguously in the report itself. The assumption of indivisibility also fails as a matter of law. The Electoral (Amendment) (Dáil Constituencies) Act~2017 defines Dublin North-West~(3) to include, in Fingal, ``those parts of the electoral divisions of Airport, Blanchardstown-Abbotstown, Dubber, The Ward and Turnapin situated south of a line drawn along the Northern Cross Route (M50)\dots'' ~\cite{ElectoralAct2017}(Fig.~\ref{fig:dubber1}), assigning parts of single EDs to different constituencies; the Electoral (Amendment) Act~2023 contains a further instance in Dublin South-Central~\cite{ElectoralAct2023}.

Two consequences follow. First, the sub-ED splits identified arise only through explicit legislative carve-outs along named physical features, such as the M50 in this instance, rather than through routine boundary drawing. They are rare, enumerable, and confined to specific constituencies, so treating the ED as the atomic node introduces an approximation that is bounded and characterisable rather than pervasive; the affected EDs can be listed and, if required, pre-split into the relevant fragments before the graph is constructed. Second, the choice of ED set should be stated explicitly wherever results are reported, because the two sets differ and a configuration admissible over one need not be admissible over the other. We adopt the CSO statistical EDs, since they are the units at which population data is released, and flag the small set of legislatively split EDs as a known limitation.


\section{Alternative Definition for Variance}\label{appendix: Alternative Definition for Variance}

This excerpt is included from our boundary submission for completeness. For a given 
constituency, an alternative variance functional, formulated in a more 
representation-central manner, is:

\[v_{\text{old}} = \frac{P/\langle P\rangle - m}{m} \qquad v_{\text{new}} = \frac{P/\langle P\rangle - m}{P/\langle P\rangle}.\]

Under $v_{\text{old}}$, rearranging for $\pm v$ at fixed $m$ gives the extremal populations:

\begin{equation}
    P_{\text{max}} = m\langle P\rangle(1+v), \qquad P_{\text{min}} = m\langle P\rangle(1-v).
\end{equation}

The misrepresentation of a constituency is the absolute discrepancy in representatives 
per constituent relative to the national average:

\begin{equation}
    \text{Misrep.}(P) = \left| \frac{m}{P} - \frac{1}{\langle P\rangle} \right|.
\end{equation}

Evaluating at the extremal populations:

\begin{equation}
    \text{Misrep.}(P_{\text{max}}) = \frac{v}{\langle P\rangle(1+v)}, \qquad 
    \text{Misrep.}(P_{\text{min}}) = \frac{v}{\langle P\rangle(1-v)}.
\end{equation}

These are unequal, so the symmetrically evaluated bounds of $v_{\text{old}}$ do not correspond to 
equal degrees of misrepresentation when viewed as representatives per constituent. Under $v_{\text{new}}$, rearranging for $\pm v$ 
similarly gives $P_{\text{max}} = m\langle P\rangle/(1-v)$ and 
$P_{\text{min}} = m\langle P\rangle/(1+v)$, and evaluating the misrepresentation:

\begin{equation}
    \text{Misrep.}(P_{\text{max}}) = \frac{v}{\langle P\rangle}, \qquad 
    \text{Misrep.}(P_{\text{min}}) = \frac{v}{\langle P\rangle}.
\end{equation}

The bounds are now symmetric in misrepresentation for any $m$-seater constituency 
and any threshold $v$, since using $P/\langle P\rangle$ (deserved representation) 
rather than $m$ (assigned representation) as the reference renders the functional 
voter-centred.

\section{Markov Chain Monte Carlo Methods}\label{app: MCMC}
Monte Carlo methods are numerical algorithms involving random sampling used to calculate expected values. For identically distributed random variables $A_1,A_2,\ldots$ with mean $\mu$, then by the strong law of large numbers,
\begin{equation}
    \label{eq:strong law}\lim_{N\to\infty}\frac{1}{N}\sum_{i=1}^NA_i = \mu.
\end{equation}
Thus, for a large enough $N$, the mean $\mu$ can be approximated by the sample mean
\begin{equation}
    \label{eq:sample mean}\hat{\mu} = \frac{1}{N}\sum_{i=1}^NA_i.
\end{equation}

A Markov chain is a sequence of ``memoryless'' states, i.e.~states which are generated based on the previous state alone. More formally, for a (homogeneous) Markov chain of random variables $A_1,A_2,\ldots$, then
\begin{equation}
    \label{eq:Markov}\begin{aligned}
    P\{A_{N+1}=a_i\,|\,A_{N}=a_j,A_{N-1}=a_{j_{N-1}},\ldots,A_0=a_{j_0}\} &= P\{A_{N+1}=a_i\,|\,A_{N}=a_j\}\\
    &= P\{A_{M+1}=a_i\,|\,A_{M}=a_j\}\,\forall\,N,M.
\end{aligned}
\end{equation}
For a probability density function $\lambda(\sigma)$ that is to be sampled from using a Markov chain, define the transition probability density function
\begin{equation}
    \label{eq:Markov transition}t(\sigma',\sigma) = P\{A_{N+1}=\sigma'\,|\,A_{N}=\sigma\}
\end{equation}
and the infinitesimal volume element $d\Gamma_{\sigma}$ in the phase space centred around state $\sigma$. $\lambda(\sigma)$ and $t(\sigma',\sigma)$ are probability density functions, and so
\begin{align}
    \label{eq:axioms}\lambda(\sigma) &\geqslant 0, & \int d\Gamma_{\sigma}\,\lambda(\sigma) &= 1, & t(\sigma',\sigma) &\geqslant 0, & \int d\Gamma_{\sigma'}\,t(\sigma',\sigma) &= 1.
\end{align}
If $t(\sigma',\sigma)$ satisfies detailed balance for $\lambda(\sigma)$, that is
\begin{equation}
    \label{eq:detailed balance}t(\sigma',\sigma)\lambda(\sigma) = t(\sigma,\sigma')\lambda(\sigma')\,\forall\,\sigma,\sigma',
\end{equation}
then $\lambda(\sigma)$ is the equilibrium distribution of $t(\sigma',\sigma)$, i.e.~choosing $t(\sigma',\sigma)$ as the transition distribution leads to a Markov chain of values sampled from $\lambda(\sigma)$ (as $\int d\Gamma_{\sigma}\,t(\sigma',\sigma)\lambda(\sigma) = \int d\Gamma_{\sigma}\,t(\sigma,\sigma')\lambda(\sigma') = \lambda(\sigma')\int d\Gamma_{\sigma}\,t(\sigma,\sigma') = \lambda(\sigma')$).
\subsection{Metropolis-Hastings Algorithm}
The Metropolis-Hastings algorithm~\cite{Hastings} operates using an ``accept/reject'' process, outlined as follows~\cite{Ross12}.

Say $\lambda(\sigma)$ is a probability density function from which values are to be sampled. Let the proposal density function $p(\sigma',\sigma)$ be a distribution corresponding to proposing a state $\sigma'$ given the current state $\sigma$, and the acceptance probability $a(\sigma',\sigma)$ be the probability that this proposed state is accepted as the next state, given by
\begin{equation}
    \label{eq:MH acceptance}a(\sigma',\sigma) = \min\!\left(\frac{p(\sigma,\sigma')\lambda(\sigma')}{p(\sigma',\sigma)\lambda(\sigma)},1\right).
\end{equation}
These expressions satisfy
\begin{equation}
    \label{eq:MH detailed balance}a(\sigma',\sigma)p(\sigma',\sigma)\lambda(\sigma) = a(\sigma,\sigma')p(\sigma,\sigma')\lambda(\sigma'),
\end{equation}
and so if the transition probability density function is defined as
\begin{equation}
    \label{eq:MH transition}t(\sigma',\sigma) = a(\sigma',\sigma)p(\sigma',\sigma),
\end{equation}
then $t(\sigma',\sigma)$ satisfies detailed balance for $\lambda(\sigma)$.

In the models proposed here, the proposed state $\sigma'$ differs to the current state $\sigma$ at a single lattice site $i$, i.e.
\begin{equation}
    \label{eq:MH proposed state}\sigma' = \left.\sigma\right|_{q_i\to q_i'},
\end{equation}
and the algorithm is repeated for each site in turn.

The choice of the proposal function $p(\sigma',\sigma)$
\subsubsection{Metropolis Algorithm}
The Metropolis algorithm~\cite{Metropolis} is a specific case of the Metropolis-Hastings algorithm, where the proposal density function is symmetric about the current state, i.e.~$p(\sigma',\sigma)=p(\sigma,\sigma')$. For $\lambda(\sigma)$ defined as in \autoref{eq:Boltzmann},
\begin{equation}
    \label{eq:Met acceptance}a(\sigma',\sigma) = \min\!\left(\frac{p(\sigma',\sigma)\lambda(\sigma')}{p(\sigma',\sigma)\lambda(\sigma)},1\right) = \min\!\left(\exp\!\left(\frac{-\Delta H(\sigma',\sigma)}{T}\right),1\right),
\end{equation}
and so only the difference $\Delta H(\sigma',\sigma)=H(\sigma')-H(\sigma)$ between the proposed and current Hamiltonian is necessary to calculate for producing the next state. Notably, if the proposed state reduces the Hamiltonian, then it is always accepted.

The simplest choice of a symmetric proposal function for updating site $i$ according to \autoref{eq:MH proposed state} is a uniform distribution over the other states, i.e.
\begin{equation}
    \label{eq:Met proposed}p(\sigma',\sigma) = \left\{\begin{array}{ll}\frac{1}{Q-1}, & q_i'\neq q_i \\
    0, & q_i'=q_i\end{array}\right..
\end{equation}
\subsubsection{Gibbs Sampler}
The Gibbs sampler method~\cite{Geman} is another type of Metropolis-Hastings algorithm, where the proposed state is generated such that it is always accepted. In particular, the proposal density function is given by~\cite{Ross12}
\begin{equation}
    \label{eq:Gibbs proposal}p(\sigma',\sigma) \propto \lambda(\sigma'),
\end{equation}
and so
\begin{equation}
    \label{eq:Gibbs acceptance}a(\sigma',\sigma) = \min\!\left(\frac{\lambda(\sigma)\lambda(\sigma')}{\lambda(\sigma')\lambda(\sigma)},1\right) = 1.
\end{equation}

The proposal function $p(\sigma',\sigma)$ for single-site updating according to \autoref{eq:MH proposed state} has the same form as $\lambda(\sigma')$, differing only by a renormalising factor. For $\lambda(\sigma)$ defined as in \autoref{eq:Boltzmann},
\begin{equation}
    \label{eq:Gibbs proposal specific}p(\sigma',\sigma) = \frac{\lambda(\sigma')}{\sum_{\sigma''}\lambda(\sigma'')} = \frac{\exp\!\left(-\frac{H(\sigma')}{T}\right)}{\sum_{\sigma''}\exp\!\left(-\frac{H(\sigma'')}{T}\right)} = \frac{\exp\!\left(-\frac{\Delta H(\sigma',\sigma)}{T}\right)}{\sum_{\sigma''}\exp\!\left(-\frac{\Delta H(\sigma'',\sigma)}{T}\right)}.
\end{equation}

\section{Contiguity Verification via Euler Characteristic}\label{EulerCharacteristic}
Verifying constituency contiguity often relies on graph traversal (BFS/DFS), which is inherently sequential and which must be repeated per constituency per optimisation step. Here we propose a matrix-based method that verifies contiguity for all constituencies simultaneously using the Euler characteristic of planar graphs, reducing the problem to sparse linear algebra that is trivially parallelisable. We leave the computational implementation to a future work. 

Let $G = (V, E, F)$ be a planar triangulation of the map, where $V$, $E$, and $F$ denote the sets of vertices (electoral districts), edges, and triangular faces respectively. Each vertex is assigned to exactly one of $m$ constituencies.

For a connected, simply connected planar region, not including the unbounded exterior surface, the Euler characteristic satisfies

\begin{equation}
\chi = v - e + f = 1.
\end{equation}

where $v$, $e$, $f$ count the vertices, edges, and faces of the induced subgraph. More generally, for a region with $k$ connected components and $h$ holes (enclosed foreign districts),

\begin{equation}
\chi = k - h.
\end{equation}

$\chi$ must equal $1$ in order to have a valid constituency with regards contiguity, meaning any other value for $\chi$ should be penalised. However $\chi=1$ is a necessary but not sufficient condition for contiguity, with failure modes that are discussed at the end of this section.
For $\chi<1$, there must be one or more enclosed EDs from one or more other constituencies. For $\chi>1$, there must be multiple fragmented, disconnected components.

\subsection{Matrix Formulation}
To efficiently calculate the Euler characteristic, first define a binary constituency assignment matrix $C \in \{0,1\}^{m \times n}$, where $C_{ij} = 1$ if district $j$ belongs to constituency $i$ and $n = |V|$. $C$ will change as the optimisation proceeds. The vertex count for each constituency is given directly by the row sums of $C$:
$$
v_i = \sum_j C_{ij}
$$

Construct a binary incidence matrix $\mathbf{B} \in \{0,1\}^{n \times |E|}$, where column $k$ has entries of $1$ at exactly the two vertices sharing edge $k$. The product $C\mathbf{B}$ yields an $m \times |E|$ matrix whose entry $(i,k)$ counts how many endpoints of edge $k$ lie in constituency $i$. An edge is interior to constituency $i$ if and only if both endpoints belong to it

$$
e_i = \sum_k \mathbf{1}\!\left[(C\mathbf{B})_{ik} = 2\right],
$$

where $\mathbf{1}[...]$ is the Iverson bracket, whereby $\mathbf{1}[(CE)_{ik}=2]$ means ``1 if that entry equals 2, else 0.''

Next, construct a weighted incidence matrix $\mathbf{F} \in \mathbb{R}^{n \times |F|}$, where column $l$ corresponding to an $n_l$-gon face has entries of $1/n_l$ at each of its $n_l$ vertices. For a triangulation, $n_l = 3$ throughout, so each column of $\mathbf{F}$ has weight $1/3$ at three vertices (we are ignoring the unbounded/outer face). A face is interior to constituency $i$ if and only if all its vertices belong to it:

$$
f_i = \sum_l \mathbf{1}\!\left[(C\mathbf{F})_{il} = 1\right]
$$

The Euler characteristic for each constituency is then
$$
\chi_i = v_i - e_i + f_i
$$

and the full vector $\boldsymbol{\chi} \in \mathbb{Z}^m$ is computed in one pass from the three sparse matrix products and summations.

The matrices $\mathbf{B}$ and $\mathbf{F}$ are sparse, and are fixed for a given map and precomputed once. Only $C$ changes during optimisation. All operations are data-parallel across constituencies and amenable to GPU acceleration.

Rather than enforcing contiguity as a hard constraint (which we found to be too rigid a constraint during optimisation), the Euler characteristic provides a geometrically meaningful penalty term:

$$
H_{C} = \sum_{i=1}^{m} |\chi_i - 1|
$$

However, we should acknowledge the presence of an exotic failure mode, which is where there are $N$ disconnected regions of a constituency, and $N-1$ holes within those regions -  the enclosed regions containing EDs from other constituencies will themselves exert an optimisation pressure to become whole, so that this is not a stable arrangement. However, one could imagine a correlated failure mode, a ``yin-yang'' type arrangement where a fragment of constituency A is in B, and vice versa, such that each contributes $\chi_A = \chi_B = 1$, so that these enclave defects are topologically reinforced. However, it may be the case that this can be remedied with a periodic BFS audit. This is a single instance of a wider class of failure modes in which $k-h=1$.

\end{document}